\documentclass[runningheads]{llncs}
\pdfoutput=1  % force pdflatex on arXiv (PDF/PNG figures)

\usepackage{eccv}

\usepackage{eccvabbrv}

\usepackage{graphicx}
\usepackage{booktabs}
\usepackage{fontawesome5}
\usepackage[accsupp]{axessibility}  % Improves PDF readability for those with disabilities.

\usepackage{hyperref}

\usepackage{orcidlink}

\usepackage{soul}

\newcommand\samethanks[1][\value{footnote}]{\footnotemark[#1]}
\def\sepappendix{0}  % combined: main paper + supplement (arXiv)
\newcommand{\suppref}[2]{\if\sepappendix1 #2\else #1\fi}
\newcommand{\mainref}[2]{\if\sepappendix1 #2\else #1\fi}
\begin{document}

\title{TrustCLIP: Learning Private Visual Features via Adversarial Reconstruction}

\titlerunning{TrustCLIP}

% Author ORCIDs and affiliations -- please double-check the flagged entries (see chat).
\author{Nikos Athanasiou\inst{1}\orcidlink{0000-0002-4722-6635} \and
Ilya A. Petrov\inst{2}\samethanks\orcidlink{0000-0002-8900-1071} \and
Angela Yao\inst{3}\orcidlink{0000-0001-7418-6141} \and
Shugao Ma\orcidlink{0000-0002-4986-2221} \and
Eric Sauser\orcidlink{0009-0007-4764-2819} \and
Edoardo Remelli\orcidlink{0000-0002-8506-9191} \and
Shreyas Hampali\orcidlink{0009-0007-3951-0832} \and
Johannes Sch\"onberger \and
Fadime Sener\orcidlink{0000-0001-5004-6005} \and
Bugra Tekin\orcidlink{0000-0001-8811-9919}}

\authorrunning{N.~Athanasiou et al.}

% Institution list -- please double-check (see chat).
\institute{Max Planck Institute for Intelligent Systems, T\"ubingen, Germany \and
University of T\"ubingen, Germany \and
National University of Singapore, Singapore}

\maketitle

\begin{figure*}[h!]
    \centering
    \includegraphics[width=0.88\textwidth]{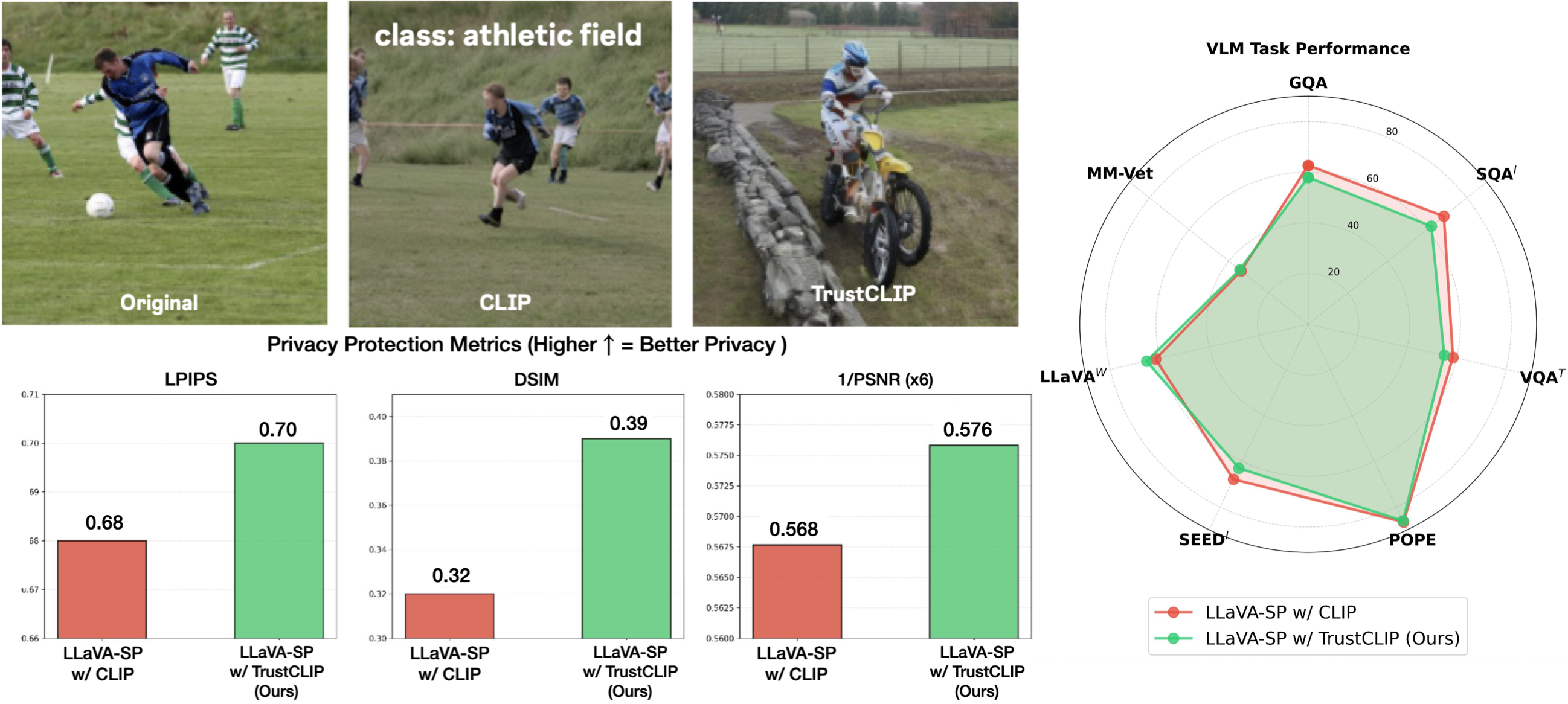}
        \caption{\textbf{TrustCLIP ensures privacy for visual understanding tasks while preserving task utility.}
(top left) Reconstructions from CLIP features reveal detailed content, whereas TrustCLIP features prevent meaningful recovery of the original image (while still maintaining class semantics \ie `semantic field').
(right) Despite the privacy enhancement, LLaVA-SP using TrustCLIP maintains competitive performance compared to the unprotected baseline across standard VLM benchmarks.
(bottom left) Privacy metrics further confirm that TrustCLIP produces substantially less reconstructible features.
}
    \label{fig:teaser}
\end{figure*}
\vspace{-1cm}
\begin{abstract}
Vision and vision–language models rely on high-level visual representations that are increasingly used across recognition, retrieval, and multimodal reasoning pipelines.
However, recent advances in generative modeling have shown that such features can often be inverted, enabling realistic reconstructions of the underlying image and raising significant privacy risks. 
We revisit this problem through the lens of reconstruction and propose \emph{TrustCLIP}, a reconstruction-driven framework that treats a feature-conditioned generator as an explicit privacy adversary.  
TrustCLIP learns a projection between encoder features and downstream modules 
that is explicitly optimized to degrade the reconstructions produced by 
generative attackers while retaining the necessary signals for downstream tasks. 
Unlike prior defenses that rely on discriminative privacy metrics, TrustCLIP 
directly optimizes against a generative reconstruction attacker, targeting a 
threat not captured by standard evaluation protocols. We demonstrate its effectiveness in both conventional classification and multimodal large language model pipelines. 
Across these settings, TrustCLIP consistently reduces the fidelity of generative inversions while maintaining downstream task performance.
Project page: \href{https://atnikos.github.io/trustclip/}{\texttt{atnikos.github.io/trustclip}}.
\end{abstract}

\section{Introduction}
\label{sec:intro}
% ==============================================================================
% =============================== (1) MOTIVATION ===============================
% ==============================================================================
Vision-language models such as CLIP~\cite{radford2021clip} have become central components in modern computer vision. The models are robust and widely used, enabling zero-shot recognition, retrieval, and other open-world reasoning tasks. 
However, this power comes with a significant privacy concern. 
Recent works have shown that CLIP features can be inverted to produce high-fidelity reconstructions of the original image~\cite{chen2025leakyclip, dosovitskiy2016inverting, struppek2022face, ye2023ip-adapter}.
Given how widely these features are deployed in modern systems, this threat is far from 
theoretical. If feature vectors collected on-device or shared with remote services are intercepted 
or misused, adversaries can reconstruct faces~\cite{wang2023privacy, shamshad2023clip2protect}, 
private home environments, medical imagery, or other sensitive content~\cite{zhou2024model, 
xiu2025caprecover}. Such reconstructions can enable re-identification, profiling, or even blackmail, 
with consequences ranging from personal harm to breaching data-protection regulations~(e.g., GDPR, 
HIPAA, the EU AI Act). 
As vision-language features are increasingly deployed in 
healthcare~\cite{sharshar2025vlmedge}, autonomous 
driving~\cite{elhenawy2025vision}, and on personal devices, the ability to 
recover raw images from stored or transmitted embeddings poses a tangible risk 
to user privacy and trust.

Mitigating this risk, however, is not straightforward. 
Downstream tasks such as classification, visual question answering, and multimodal reasoning depend 
on the rich semantic content encoded in CLIP features. Naively suppressing information---for example by adding noise or quantizing the representation---degrades these features indiscriminately, 
harming both the reconstruction-relevant and the task-relevant components alike. 
The core difficulty is that the visual information exploited by a reconstruction attacker is deeply 
entangled with the information needed to solve downstream tasks: 
both rely on the same high-dimensional feature space produced 
by visual encoders. 
Moreover, most existing privacy defenses are evaluated against \emph{discriminative} metrics---such 
as the accuracy of an attribute classifier on the protected features---which fail to capture 
\emph{generative} leakage, where a diffusion-based attacker reconstructs visually coherent images 
that reveal sensitive content even when re-identification accuracy is low. 
Any privacy mechanism must therefore be selective, suppressing the components that enable faithful 
image recovery while preserving those that carry task-critical semantics.
This raises a fundamental question: \emph{how can we preserve the utility of CLIP features for downstream tasks, while mitigating the privacy risks that arise from their misuse?}

\begin{figure}[t]
    \centering
    \includegraphics[width=\columnwidth]{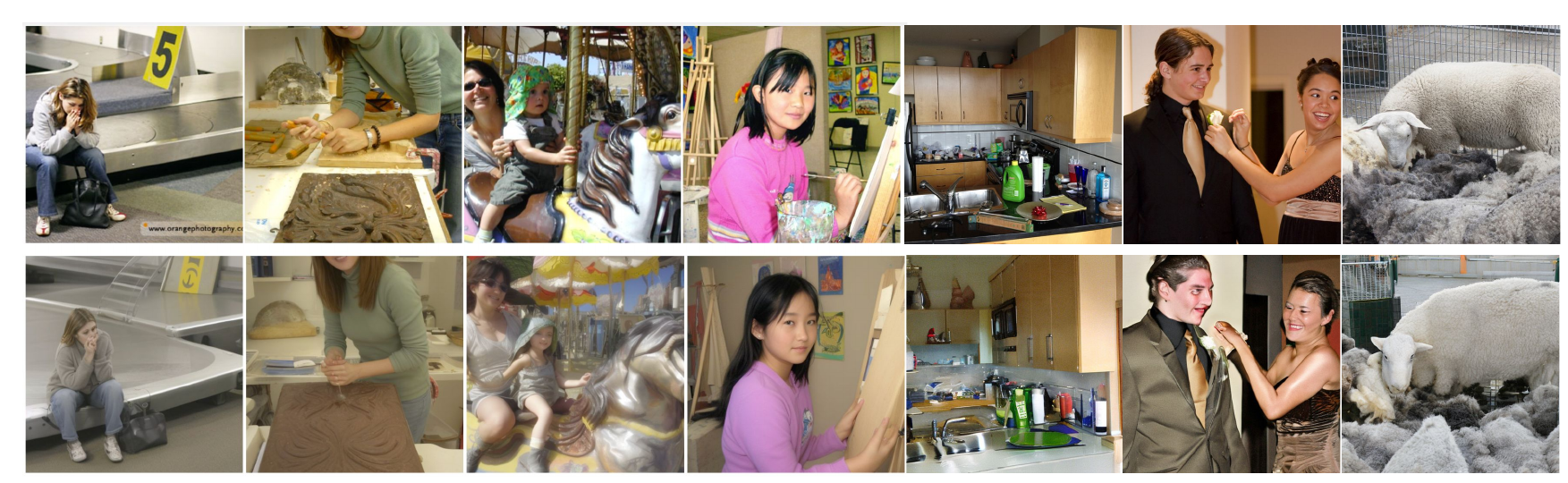}
    \vspace{-3mm}
       \caption{\textbf{CLIP features reveal privacy information.}
Top: original images. Bottom: IP-Adapter reconstructions from frozen CLIP Vision features. Outputs closely match the originals, preserving layout, pose, and fine-grained semantics, including sensitive cues such as perceived gender, skin tone, age (children vs.\ adults), facial affect (e.g., desperation vs.\ smile), and detailed indoor/outdoor context. %—illustrating how much private information CLIP features can reveal.
}
    \label{fig:motivation}
\end{figure}

% ==============================================================================
% ======================= (2) WHAT PREVIOUS WORK HAS DONE ====================== 
% ==============================================================================
Prior work has explored privacy-preserving representations through adversarial projection~\cite{dave2022spact}, video obfuscation~\cite{ilic2024selective}, and secure aggregation~\cite{truex2019hybrid}, yet existing defenses either sacrifice task performance or only partially reduce reconstruction fidelity, leaving an unsatisfactory privacy--utility trade-off. 
In this work, we propose \emph{TrustCLIP}, a framework for learning privacy-preserving visual representations against generative inversion attacks without  sacrificing task-specific performance.
We consider a realistic threat model in which an adversary has access to intermediate vision features\footnote{For example, in client-server inference where a device sends CLIP 
features to a cloud VLM or in retrieval systems that cache embeddings.} produced by a frozen encoder and to a powerful image-conditioned generative model that can synthesize images directly from these features~(see Fig.~\ref{fig:motivation}). Rather than reasoning about privacy only through proxy metrics on latent space, we explicitly model this reconstruction pathway and treat \emph{invertibility under a generative attacker} as the primary privacy risk.
Our approach inserts a lightweight, adversarially-trained projection between the frozen CLIP encoder 
and different downstream networks using its features. The projection is optimized under two competing 
objectives: a \emph{task loss} that preserves downstream accuracy, and a \emph{reconstruction loss} 
that penalizes the fidelity of images recovered by a generative attacker conditioned on the 
projected features. This adversarial formulation forces the projection to discover which components 
of the representation are critical for reconstruction versus those that are critical for the task, 
and to selectively suppress the former.
The premise underlying this approach is that downstream tasks and reconstruction attacks do not 
depend on the \emph{same} information within CLIP features to the same extent.
Tasks such as classification and visual question answering primarily depend on high-level semantic structure---object categories, 
spatial layout, and coarse scene understanding---whereas faithful image reconstruction additionally requires 
instance-specific detail: fine textures, skin tones, facial geometry, and identity-revealing cues. 
Because these informational demands are not fully overlapping, a learned projection can degrade 
reconstruction fidelity without proportionally harming task performance. 
Our experiments confirm this separation (Fig.~\ref{fig:teaser}): TrustCLIP 
features preserve scene-level semantics (Top-1 accuracy within 0.5\% of the 
unprotected baseline) while obliterating the fine-grained detail that enables 
identity recovery (DSIM improves 2.5$\times$). 
The projection itself is a lightweight MLP that integrates naturally 
into existing pipelines---we demonstrate its effectiveness in both image classification and 
vision--language models, showing that the adversarial training framework generalizes across diverse 
downstream architectures.

We instantiate our framework using IP-Adapter~\cite{ye2023ip-adapter}, a diffusion-based generator 
conditioned on CLIP features, which represents one of the strongest publicly available 
reconstruction attacks for vision--language embeddings. Although we focus on CLIP and IP-Adapter, 
the adversarial training formulation is not specific to this combination and can be applied to other 
encoders or generative attackers. 
Our contributions are as follows:
\textbf{(i)} We identify generative leakage---the ability of a diffusion-based attacker to 
reconstruct privacy-revealing images from encoder features---as a distinct and underexplored threat 
not captured by standard discriminative privacy metrics, and introduce an evaluation framework that directly measures it.
\textbf{(ii)} We propose a lightweight, adversarially-trained projection that explicitly reduces reconstruction fidelity under the attacker while maintaining task utility, yielding improved privacy--utility trade-offs.
\textbf{(iii)} We construct privacy-preserving variants of both conventional image classification models and multimodal large language models (MLLMs), demonstrating that the projection layer integrates naturally into these architectures and supports privacy across diverse downstream tasks.
\textbf{(iv)}  We perform extensive evaluations across privacy metrics (LPIPS, DreamSim) and downstream benchmarks in classification and vision–language modeling, demonstrating that the proposed method significantly improves privacy preservation while retaining competitive performance.

% ~\AY{introduction is too short - i think need more to support the understanding of the method, and why your are able to overcome the utiltiy / privacy trade-off problem.  currently, there are 2 paragraphs of motivation about why privacy is necessary, one paragraph on the setup of the intermediate features, and then only 1 method explaining your contribution which is second last paragraph, before jumping to last paragraph on summary of contributions. Needs more substance on the second-last paragraph to clearly articulate what your novelty is and why it overcomes the issues you mention previously.}

\vspace{-0.5cm}
\section{Related Work}

\noindent\textbf{Feature inversion \& privacy leakage.}
Deep representations retain enough information to reconstruct
input images, whether by optimizing pixels to match
activations~\cite{mahendran2015understanding}, learning
feature-to-image decoders~\cite{dosovitskiy2016inverting}, or
exploiting models to reveal training-data
prototypes~\cite{fredrikson2015modelinversion,carlini2023extracting}.
Powerful generative priors encoded in GANs and diffusion
models dramatically improve reconstruction
fidelity~\cite{dhariwal2021diffusion,hintersdorf2024does,meng2021sdedit,chen2025enhancing}.
CLIP~\cite{radford2021clip,ilharco2021openclip} is especially
vulnerable: its patch tokens encode rich semantics while staying closely aligned to natural-image statistics through web-scale contrastive pretraining.  This property makes CLIP embeddings %significantly 
more reconstructible than CNN or self-supervised
descriptors~\cite{chen2025leakyclip,kazemi2024we}.

\noindent\textbf{Diffusion adapters \& CLIP-conditioned generation.}
Diffusion models~\cite{ho2020ddpm,song2020score,Rombach_2022_CVPR}
are the dominant backbone for high-fidelity synthesis.
Lightweight adapters such as ControlNet~\cite{zhang2023controlnet},
T2I-Adapter~\cite{mou2023t2iadapter},
IP-Adapter~\cite{ye2023ip-adapter} condition the denoising
process on spatial or feature-level guidance while keeping the backbone frozen. IP-Adapter maps CLIP features into the
text-conditioning space, letting a frozen diffusion U-Net
reconstruct rich semantic and perceptual detail from visual
tokens alone. Because both CLIP and the diffusion backbone are
trained on natural image--text alignments, their latent spaces
are inherently compatible---making this a strong yet practical
adversary, as all components are publicly released. We exploit
this property as our white-box attack.

\noindent\textbf{Privacy-preserving representations.}
Approaches to suppress sensitive information while preserving
task utility include adversarial
disentanglement~\cite{edwards2015censoring,kim2019learning},
information bottlenecks~\cite{alemi2016deep}. 
In video, SPAct~\cite{dave2022spact} and
STPrivacy~\cite{li2023stprivacy} adversarially train encoders
that maintain action recognition while concealing identity;
hardware solutions such as coded-aperture
recognition~\cite{wang2019privacy} offer non-invertible capture;
and Ilic et al.~\cite{ilic2024selective} use optical flow with
DINO features for motion-consistent obfuscation.
DP-CLIP~\cite{huang2023safeguarding} applies differential
privacy at the batch level during contrastive training, offering
formal guarantees at the cost of accuracy.
A key limitation of most defenses is their reliance on
\emph{proxy} privacy metrics---attribute-classifier accuracy or
mutual-information estimates---which capture
\emph{discriminative} leakage but miss \emph{generative}
leakage, where a diffusion-based adversary reconstructs
privacy-revealing images despite low re-identification accuracy.
TrustCLIP addresses this gap: it is, to our knowledge, the first defense that directly 
optimizes against a generative reconstruction attacker, treating the fidelity 
of diffusion-based inversion---rather than attribute-classifier accuracy---as the primary 
privacy objective. Our framework explicitly evaluates privacy under such a
generative threat. 

\noindent\textbf{Inversion-resistant descriptors \& scene-level privacy.}
In retrieval and localization,
NinjaDesc~\cite{ng2022ninjadesc} adversarially trains
descriptors that resist decoder-based reconstruction, while 
Adversarial Affine Subspace
Embeddings~\cite{dusmanu2021privacy} obfuscate features while
preserving geometric matching. 
At the scene level, geometric representations also leak appearance: SfM point clouds can be 
inverted to recover images~\cite{pittaluga2019revealing}, and subsequent work has proposed 
mitigations through geometric substitutions~\cite{speciale2019privacy}, uncalibrated 
localization~\cite{geppert2021privloc}, and ray clouds~\cite{moon2024raycloud}.
These methods target descriptor-level or geometric reconstruction.
In contrast, we address a complementary and increasingly critical issue: the inversion 
of language-aligned vision features---specifically CLIP tokens---through diffusion-based 
generation, a threat that grows in relevance as such features become standard 
between vision encoders and downstream models in modern VLMs.

\section{Method}
\label{sec:method}

%% ============================================================
%% §3.1 OVERVIEW
%% ============================================================
% \subsection{Overview}
% \label{subsec:overview}

\begin{figure*}[t]
    \centering
    %% ---- REPLACE with your redesigned architecture figure ----
    \includegraphics[width=\textwidth]{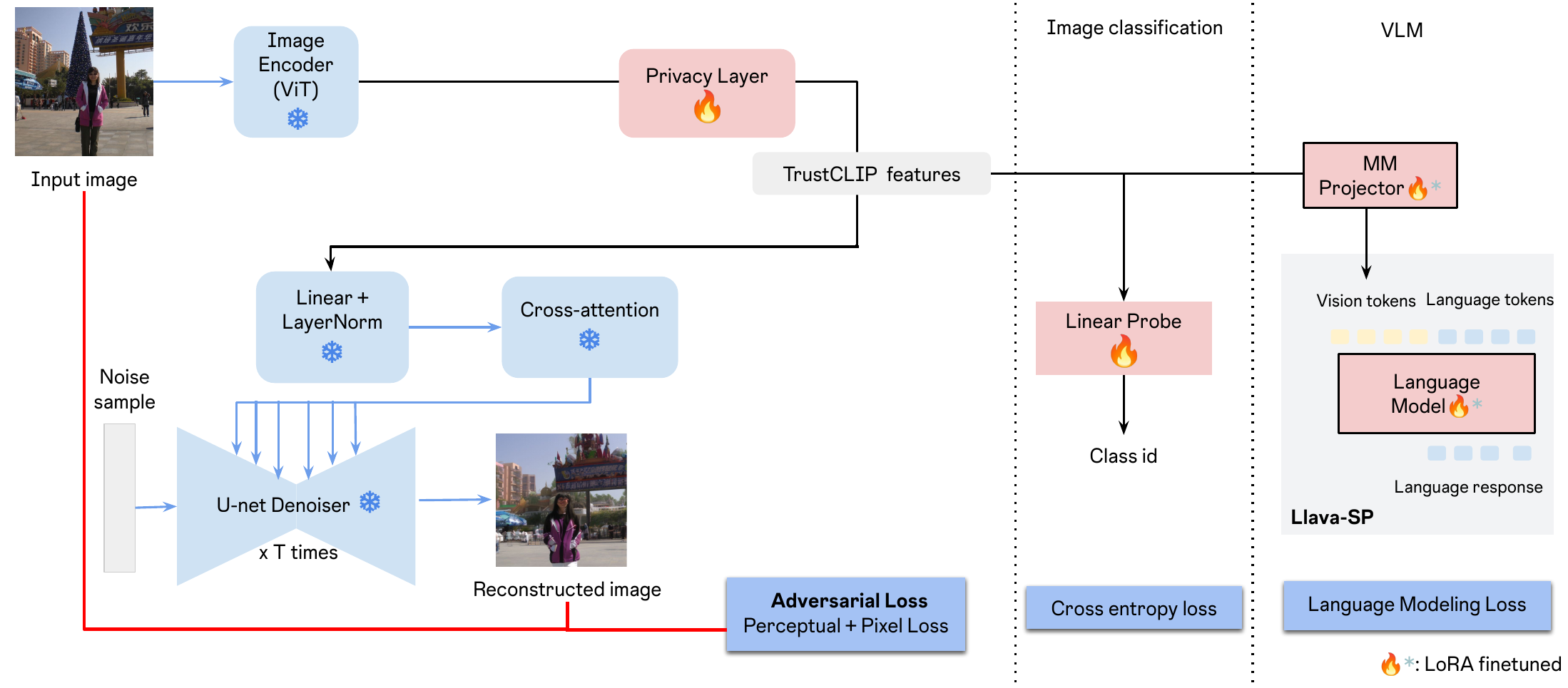}
    \caption{\textbf{Overview of TrustCLIP.}
    A frozen vision encoder extracts feature tokens from an input image.
    The privacy projection $P_\theta$ transforms these features before they are consumed by any downstream module.
    During training (\textcolor{red}{red path}), a frozen generative attacker $G_\phi$ attempts to reconstruct the original image from the projected features; the reconstruction loss gradient is backpropagated through the attacker to update $P_\theta$.
    The task heads (image classification/VLM) simultaneously optimize task performance.
    At deployment, only the non-red path is active: the attacker is discarded, and $P_\theta$ functions as a lightweight, drop-in privacy layer.
    \faSnowflake\ frozen;\, \faFire\ trainable.}
    % \snowflake: frozen;\, \fire: trainable.}
    \label{fig:overview}
\end{figure*}

Figure~\ref{fig:overview} illustrates the TrustCLIP framework.
A frozen vision encoder $f_v$ (e.g.\ CLIP ViT-L/14) maps an input image $x$ to a sequence of feature tokens $z = f_v(x) \in \mathbb{R}^{T \times D}$.
A lightweight privacy projection $P_\theta$ transforms these tokens into $\tilde{z} = P_\theta(z)$ before they reach any downstream consumer.
Two modules compete during training: a \emph{task head} $h_{\text{task}}$ that produces predictions $\hat{y} = h_{\text{task}}(\tilde{z})$ under a task-specific loss (\S\ref{subsec:adv-objective}), and a generative \emph{attacker} $G_\phi$ that attempts to reconstruct the original image $\tilde{x} = G_\phi(\tilde{z})$ (\S\ref{subsec:attacker}).
The projection $P_\theta$ and task head are jointly optimized to minimize task loss while maximizing reconstruction error, so that $P_\theta$ learns to suppress information that enables image recovery while preserving information that supports the downstream task.
At deployment, only the green path in Figure~\ref{fig:overview} is used: the attacker is discarded, and $P_\theta$ serves as a drop-in privacy layer that adds negligible overhead and requires no architectural modification to the encoder or downstream model.
We emphasize that, while $P_\theta$ itself is lightweight, the downstream task head is fine-tuned jointly with it---a linear probe for classification and LoRA adapters for the VLM (\S\ref{subsec:adv-objective})---so the contribution is a privacy layer that requires no change to the network \emph{architecture}, rather than one that leaves the downstream model entirely untouched.

%% ============================================================
%% §3.2 THREAT MODEL
%% ============================================================

\paragraph{Threat Model.}
\label{subsec:threat-model}
We consider a setting in which an adversary gains access to intermediate vision features---for example, CLIP embeddings transmitted from a client device to a cloud-hosted VLM, cached in a retrieval database, or exposed through a multi-tenant storage failure.
Given these features, the adversary attempts to reconstruct the original image using a strong generative model: concretely, an IP-Adapter~\cite{ye2023ip-adapter} conditioned on Stable Diffusion~\cite{Rombach_2022_CVPR}.
To ensure a rigorous evaluation, each attacker is an IP-Adapter \emph{finetuned on the feature distribution it attacks}: the attacker for unprotected CLIP features is trained on CLIP features, and the attacker for TrustCLIP features is trained on TrustCLIP features.
This gives every method its own strongest possible adversary.
The goal of TrustCLIP is to learn a projection $P_\theta$ that degrades the fidelity of images reconstructed from the projected features, while preserving the semantic content needed for downstream tasks.
\S\ref{subsec:premise} argues why this selective suppression is feasible.

\subsection{Privacy-Preserving Projection}
\label{subsec:projection}

\paragraph{Why selective suppression is feasible.}
\label{subsec:premise}
The success of TrustCLIP rests on a structural property of CLIP 
representations: downstream tasks primarily require high-level semantic 
content---object categories, spatial layout, coarse scene structure---whereas 
pixel-accurate reconstruction additionally requires instance-specific detail 
such as textures, skin tones, and facial geometry. Because these informational 
demands do not fully overlap, a learned projection can suppress the latter 
while preserving the former. Our experiments confirm this empirically: 
TrustCLIP features retain scene-level accuracy within 0.5\% of the baseline 
while degrading reconstruction fidelity by 2.5$\times$ in DSIM 
(\S\ref{subsec:main-results}). The naive baselines in 
Tab.~\ref{tab:naive_baselines} further validate this premise: Gaussian noise 
cannot achieve this separation because it perturbs all feature dimensions 
indiscriminately, whereas TrustCLIP's adversarial training learns to target 
only the reconstruction-critical components. 
We require $P_\theta : \mathbb{R}^{T \times D} \rightarrow \mathbb{R}^{T \times D}$ transforms encoder features into $\tilde{z} = P_\theta(z)$ such that reconstructible information is suppressed while task-relevant structure is maintained.
% \paragraph{Design requirements.}
We require $P_\theta$ to satisfy three constraints:
(i)~\emph{Model-agnostic}: it operates purely on the feature sequence, independent of the specific encoder or task head.
(ii)~\emph{Lightweight}: it adds negligible runtime overhead compared to the backbone encoder.
(iii)~\emph{Non-destructive at initialization}: at the start of training, it preserves the original representation to avoid cold-start degradation of utility.
% \paragraph{Architecture.}
We implement $P_\theta$ as a shallow, token-wise network with a residual connection, $P_\theta(z) = z + f_\theta(z)$, 
% \begin{equation}
% \label{eq:projection}
% \end{equation}
where $f_\theta$ is a small MLP applied independently to each token (optionally with layer normalization).
The last linear layer of $f_\theta$ is initialized to zero, so $P_\theta$ starts as the identity mapping.
This yields a smooth \emph{privacy--utility knob}: as $\lambda_{\text{rec}}$ 
increases (\S\ref{subsec:adv-objective}), $P_\theta$ gradually deviates from 
identity to suppress the reconstructible components while the task loss protects 
the semantic subspace. 
Although our experiments instantiate $f_v$ with CLIP, the projection itself is agnostic to the backbone and can be placed in front of any consumer of tokenized vision features, including VLMs and non-contrastive encoders.

%% ============================================================
%% §3.5 GENERATIVE ATTACKER
%% ============================================================
\subsection{Generative Attacker}
\label{subsec:attacker}

The attacker $G_\phi : \mathbb{R}^{T \times D} \rightarrow \mathcal{X}$ is a generative model that takes feature tokens as conditioning and produces a reconstruction $\tilde{x} = G_\phi(\tilde{z})$.
In principle, $G_\phi$ can be any feature-conditioned image decoder: a diffusion model with cross-attention adapters, a GAN with feature conditioning, or a latent diffusion model.

\paragraph{Training-time attacker (fixed).}
In our experiments, $G_\phi$ is an IP-Adapter module built on a Stable Diffusion backbone~\cite{ye2023ip-adapter, Rombach_2022_CVPR}.
The projected features $\tilde{z}$ are converted into conditioning tokens for the diffusion U-Net via a lightweight adapter.
This model is pre-trained on unprotected CLIP features and remains \emph{frozen} while $P_\theta$ is optimized.
Freezing the attacker during projection training provides a stable adversarial signal and avoids the instabilities of alternating min-max optimization~\cite{Mescheder2018WhichTM}.

\paragraph{Evaluation-time attacker (adaptive).}
To evaluate robustness under a worst-case adversary, we conduct a second stage after $P_\theta$ converges: a new generative model $G_{\phi'}$ is retrained \emph{directly} on the protected features $\tilde{z}$ produced by the converged projection $P_\theta^\star$:
\begin{equation}
\label{eq:adaptive-attacker}
    \phi' = \operatorname*{arg\,min}_\phi \; \mathbb{E}_{x \sim \mathcal{D}} \big[ \mathcal{L}_{\text{rec}}(x,\; G_\phi(P_{\theta^\star}(f_v(x)))) \big].
\end{equation}
This attacker has full knowledge of the defense---access to the projected feature distribution and freedom to optimize its decoder---representing the adaptive adversary of \S\ref{subsec:threat-model}.

IP-Adapter combined with Stable Diffusion represents the current frontier for 
feature-conditioned image generation: the diffusion backbone is trained on 
billions of images, and the adapter is optimized on millions of image--feature 
pairs, making it a near-optimal decoder for CLIP features. All components are 
publicly released, so any motivated adversary can deploy this attack. 
By evaluating against both the fixed attacker and a stronger adaptive variant 
retrained on the protected feature distribution 
(Eq.~\ref{eq:adaptive-attacker}), we test TrustCLIP under a realistic 
worst-case scenario. Although we focus on this specific attacker, the 
projection reduces the information available in $\tilde{z}$ for 
\emph{any} reconstruction method: by the data processing inequality, no 
downstream function of $\tilde{z}$ can recover more about $x$ than 
$\tilde{z}$ itself contains.

%% ============================================================
%% §3.6 TRAINING OBJECTIVE
%% ============================================================
\subsection{Training Objective}
\label{subsec:adv-objective}

Given an input $x$ with task label $y$, encoder features $z = f_v(x)$, and projected features $\tilde{z} = P_\theta(z)$, the task head produces predictions $\hat{y} = h_{\text{task}}(\tilde{z})$ and the frozen attacker produces a reconstruction $\tilde{x} = G_\phi(\tilde{z})$.
The reconstruction loss combines pixel-level and perceptual distances: $
    \mathcal{L}_{\text{rec}}(x, \tilde{x}) =
    \alpha \, \|x - \tilde{x}\|_p +
    (1-\alpha)\,\text{LPIPS}(x, \tilde{x}) $,
where $\|\cdot\|_p$ is an $\ell_p$ pixel distance ($p \in \{1,2\}$) and LPIPS~\cite{zhang2018perceptual} provides perceptual similarity.
The full objective jointly optimizes the projection $P_\theta$ (parameters $\theta$) and the task head $h_{\text{task}}$ (parameters $\psi$):
\begin{equation}
\label{eq:adv-objective}
    \mathcal{L}(\theta,\psi) =
    \mathbb{E}_{x,y}\!\left[
        \mathcal{L}_{\text{task}}(h_\psi(\tilde{z}),\, y)
        \;-\; \lambda_{\text{rec}}\,\mathcal{L}_{\text{rec}}(x,\, G_\phi(\tilde{z}))
    \right].
\end{equation}
The first term drives task performance; the second, with a negative sign, encourages $P_\theta$ to \emph{increase} reconstruction error under the fixed attacker.
The scalar $\lambda_{\text{rec}} \geq 0$ controls the privacy--utility trade-off.
In practice, we differentiate through the frozen attacker and backpropagate the negative reconstruction gradient to $\theta$.

% \paragraph{Task heads.}
% The framework supports arbitrary downstream consumers. 
% For \emph{image classification}, $h_{\text{task}}$ is a linear probe on pooled features, trained with cross-entropy.
% For \emph{vision-language modeling}, $P_\theta$ is inserted before the VLM's visual adapter (e.g., LLaVA-SP's multimodal projector~\cite{llava-sp}), and $\mathcal{L}_{\text{task}}$ is the standard language modeling loss.
% In both cases, the encoder $f_v$ remains frozen and only $P_\theta$ and $h_{\text{task}}$ are updated.
% Detailed task-head architectures and training hyperparameters are provided in the supplementary material.

% \paragraph{Why joint optimization?}
% Joint end-to-end training is a deliberate design choice.
% A two-stage alternative---first optimizing $P_\theta$ for privacy alone, then 
% training $h_{\text{task}}$ on frozen projected features---would corrupt 
% features indiscriminately, because without task-loss gradients the privacy 
% objective has no signal about which dimensions to preserve.
% Joint optimization lets the competing gradients from $\mathcal{L}_{\text{task}}$ 
% and $\mathcal{L}_{\text{rec}}$ reach an equilibrium where reconstructible 
% information is suppressed and task-relevant structure is maintained.
% This is further supported by our choice of identity initialization 
% (\S\ref{subsec:projection}): starting from the original CLIP representation 
% ensures that both objectives co-adapt gradually, rather than forcing the 
% projection to recover useful structure after an unconstrained privacy-only 
% phase.

\paragraph{Task heads.}
The framework supports arbitrary downstream tasks. For \emph{image classification}, $h_{\text{task}}$ is a linear probe on pooled features trained with cross-entropy. For \emph{vision--language modeling}, $P_\theta$ is placed before the VLM visual adapter (e.g., LLaVA-SP’s multimodal projector~\cite{llava-sp}) and $\mathcal{L}_{\text{task}}$ is the standard language-modeling loss. In both settings, $f_v$ is frozen; only $P_\theta$ and $h_{\text{task}}$ are updated. Full task-head details and hyperparameters are provided in \suppref{Appendix~\ref{sec:supp_implementation}}{the supplementary material}.

\paragraph{Why joint optimization?}
We train end-to-end by design. A two-stage baseline---optimize $P_\theta$ for privacy, then train $h_{\text{task}}$ on frozen projected features---would corrupt features indiscriminately, since the privacy objective alone provides no signal about what to preserve. Joint optimization balances gradients from $\mathcal{L}_{\text{task}}$ and $\mathcal{L}_{\text{rec}}$, suppressing reconstructible information while retaining task-relevant structure. Identity initialization (\S\ref{subsec:projection}) further stabilizes this process: starting from the original CLIP representation lets both objectives co-adapt gradually, rather than requiring the projection to relearn useful structure after an unconstrained privacy-only phase.

\paragraph{Identity initialization and warm-up.}
The projection starts as the identity ($f_\theta$ initialized to zero), so training begins from the original CLIP features.
For classification, $\lambda_{\text{rec}}$ is constant from the start.
For VLM training, we adopt a gradual warm-up: $\lambda_{\text{rec}}$ is set to zero for the first $N$ steps while the task head adapts to the feature space, and is then linearly increased to its target value. More details in \suppref{Appendix~\ref{sec:supp_vlm_training}}{the supplementary material}.
% This schedule prevents early-stage privacy gradients from destabilizing the large VLM before the task head has calibrated.
% Combined with identity initialization, it provides a smooth and controllable transition from full utility toward the desired privacy--utility operating point.

\section{Experiments}
\label{sec:experiments}

We evaluate TrustCLIP on image classification (SUN397) and multimodal 
reasoning (LLaVA-SP), in classification and VLM benchmarks. All 
qualitative results use the adaptive attacker unless noted otherwise.

% ==============================================================
% §4.1 SETUP
% SOURCE: existing main paper, tightened
% ==============================================================
\subsection{Setup}
\label{subsec:setup}

\noindent\textbf{Backbones.}
For classification, we attach a linear probe to the frozen CLIP ViT-L/14 
encoder. For multimodal reasoning, we integrate TrustCLIP into 
LLaVA-SP~\cite{llava-sp}, yielding \emph{TrustLLaVA}. The VLM uses 
CLIP-ViT-L/14@336 as the vision encoder and Vicuna-1.5-7B as the LLM, 
trained with 558K image--text pairs and 665K instruction-following examples
in a two-stage pre-train + LoRA regime.
For a fair comparison, TrustLLaVA follows the \emph{exact} LLaVA-SP pipeline
(identical training data and LoRA configuration), differing only by $P_\theta$ and
the adversarial loss; the LLaVA-SP row in Tab.~\ref{tab:vlm_unified} is thus its
matched, in-distribution-finetuned counterpart, not an off-the-shelf VLM.

\noindent\textbf{Datasets.}
For classification: SUN397~\cite{Xiao2010SUNDL} with standard 
train/val/test splits; images resized to $512{\times}512$ for reconstruction 
and $224{\times}224$ for feature extraction.
For VLMs: the standard benchmark suite used in LLaVA-SP~\cite{llava-sp}.

% ---VQAv2~\cite{goyal2017making}, TextVQA~\cite{singh2019towards}, ScienceQA-Image~\cite{lu2022learn}, GQA~\cite{hudson2019gqa}, MM-Vet~\cite{yu2023mmvet}, LLaVA-Bench~\cite{liu2023visualinstruction},\allowbreak\ MME-Perception~\cite{fu2023mme}, SEED-Bench~\cite{li2024seedbench}, and POPE~\cite{li2023pope}.

\noindent\textbf{Attacker.}
Our generative attacker is an IP-Adapter~\cite{ye2023ip-adapter} module 
built on Stable Diffusion v1.5~\cite{Rombach_2022_CVPR}. Projected features 
are mapped to conditioning tokens for the diffusion U-Net, which 
reconstructs images using DDIM with 20 inference steps. Random seeds are 
fixed so that pre-/post-projection comparisons share identical noise. 
Attacker training details and capacity analysis are provided in
\suppref{Appendix~\ref{sec:sup_attacker_training}}{the supplementary material}.

\noindent\textbf{Metrics.}
\emph{Privacy:} PSNR, SSIM (pixel fidelity), LPIPS~\cite{zhang2018perceptual}, 
DreamSim (DSIM)~\cite{fu2023dreamsim} (perceptual/semantic distance). Lower 
PSNR/SSIM and higher LPIPS/DSIM indicate stronger privacy.
\emph{Utility:} Top-1, Top-5, and mean-class accuracy for classification; 
official benchmark metrics for VLMs.

\noindent\textbf{Baselines.}
We compare TrustCLIP to the unprotected baseline in which CLIP features are passed directly to both the attacker and downstream heads (\emph{CLIP+IP-Adapter}), illustrating the privacy risk in current systems.
For VLMs, we report alongside Qwen-VL~\cite{bai2023qwen}, Qwen-VL-Chat~\cite{bai2023qwen}, LLaVA-1.5~\cite{liu2024improved}, and LLaVA-SP~\cite{llava-sp}.
To our knowledge, TrustCLIP is the first method to adversarially modify language-aligned vision features using a generative attacker.
Prior methods such as ~\cite{ng2022ninjadesc} and DP-CLIP~\cite{huang2023safeguarding} target fundamentally different settings---sparse local descriptors for geometric matching, and training-time differential privacy for the CLIP pretraining corpus, respectively---and are therefore not directly comparable.
These differences are definitional: DP-CLIP~\cite{huang2023safeguarding}
bounds \emph{training-set membership} rather than inference-time reconstruction,
while NinjaDesc~\cite{ng2022ninjadesc} and SPAct~\cite{dave2022spact} defend a
single task against \emph{discriminative} attackers---whereas we defend
foundation-model features serving many tasks against a \emph{generative} inverter.
Porting them to dense CLIP tokens under a generative attacker would change each
method beyond recognition, so the matched-$\ell_2$ Gaussian control
(\S\ref{subsec:main-results}) instead gives the apples-to-apples comparison they
cannot.

\noindent\textbf{Implementation details.}
During training, the CLIP encoder and all attacker components are frozen; 
only $P_\theta$ and the task head are updated. At evaluation time, we additionally 
test against an adaptive attacker retrained on the protected features 
(\S\ref{subsec:main-results}).
$P_\theta$ is initialized to identity (\S\ref{subsec:projection}), ensuring training begins from the original CLIP representation.
We train with AdamW using fixed random seeds.
Full hyperparameters are provided in \suppref{Appendix~\ref{sec:supp_implementation}}{the supplementary material}.

% ==============================================================
% §4.2 DOWNSTREAM TASK PERFORMANCE
% CHANGE FROM CVPR: Honest framing — no "SOTA" claim.
% SOURCE: utility columns of Table 1, Table 3.
% ==============================================================
\subsection{Main Results}
\label{subsec:main-results}

\paragraph{Classification.}
Tab.~\ref{tab:utility_privacy} reports both task accuracy and reconstruction 
metrics on SUN397. Across all TrustCLIP 
configurations, Top-1 accuracy remains within 0.5\% of the unprotected CLIP 
baseline (83.4\%), with the best setting (L2+LPIPS, 
$\lambda_{\text{rec}}{=}0.25$) achieving 82.9\%. Simultaneously, PSNR drops 
from 13.58 to 10.47--10.69 (a 21--23\% reduction), SSIM from 0.288 to 
0.187--0.204, and DSIM improves 2.5$\times$. The unprotected baseline 
confirms the severity of the privacy risk: despite strong accuracy, its 
reconstructions remain highly faithful.
\begin{table*}[t]
\centering
\footnotesize
\caption{
\textbf{Privacy-utility trade-offs on image classification.} Evaluated using classification and reconstruction metrics for TrustCLIP variants and the CLIP using IP-Adapter for adversarial attack. Higher values indicate better classification performance; lower PSNR/SSIM and higher LPIPS/DSIM reflect stronger privacy preservation.
}
\label{tab:utility_privacy}
\resizebox{\textwidth}{!}{%
\begin{tabular}{l l l r r r r r r r r}
\toprule
& & &
& \multicolumn{3}{c}{\textbf{Classification metrics}}
& \multicolumn{4}{c}{\textbf{Reconstruction errors (mean $\pm$ std)}} \\
\cmidrule(lr){5-7} \cmidrule(lr){8-11}
Setting & Pixel & Perc. & $\lambda_{\mathrm{rec}}$ &
Top-1 & Top-5 & Mean Cl. &
PSNR $\downarrow$ &
SSIM $\downarrow$ &
LPIPS $\uparrow$ &
DSIM $\uparrow$ \\
\midrule
CLIP 
& --- & --- & ---
& 83.36 & 97.76 & 80.46
& $13.581 \pm 2.203$
& $0.288 \pm 0.147$
& $0.522 \pm 0.066$
& $0.215 \pm 0.055$ \\
\midrule

TrustCLIP & L1 & --- & 0.25
& 82.662 & 97.361 & 79.871
& $10.555 \pm 1.608$
& $0.191 \pm 0.102$
& $0.704 \pm 0.051$
& $0.522 \pm 0.098$ \\

TrustCLIP & L1 & LPIPS & 0.25
& 78.547 & 95.899 & 72.376
& $10.639 \pm 1.659$
& $0.204 \pm 0.109$
& $0.702 \pm 0.054$
& $0.535 \pm 0.098$ \\

TrustCLIP & L2 & --- & 0.25
& 82.487 & 97.411 & 79.621
& $\underline{10.527 \pm 1.576}$
& $\mathbf{0.187 \pm 0.097}$
& $0.702 \pm 0.051$
& $0.514 \pm 0.097$ \\

TrustCLIP & L2 & LPIPS & 0.25
& $\mathbf{82.915}$ & $\underline{97.582}$ & 79.270
& $10.687 \pm 1.655$
& $0.203 \pm 0.109$
& $0.699 \pm 0.053$
& $0.514 \pm 0.097$ \\

TrustCLIP & L2 & --- & 0.5
& 82.653 & 97.434 & $\underline{79.892}$
& $10.681 \pm 1.622$
& $0.195 \pm 0.101$
& $0.701 \pm 0.049$
& $0.523 \pm 0.099$ \\

TrustCLIP & L2 & LPIPS & 0.5
& 81.411 & 97.140 & 76.771
& $10.685 \pm 1.661$
& $0.200 \pm 0.108$
& $0.697 \pm 0.052$
& $0.510 \pm 0.095$ \\

TrustCLIP & L2 & --- & 1
& $\underline{82.869}$ & 97.517 & $\mathbf{80.068}$
& $\mathbf{10.465 \pm 1.616}$
& $\underline{0.188 \pm 0.100}$
& $\underline{0.714 \pm 0.051}$
& $\underline{0.556 \pm 0.110}$ \\

TrustCLIP & L2 & LPIPS & 1
& 82.763 & $\mathbf{97.669}$ & 79.070
& $10.617 \pm 1.590$
& $0.203 \pm 0.093$
& $\mathbf{0.776 \pm 0.061}$
& $\mathbf{0.799 \pm 0.068}$ \\

\bottomrule
\end{tabular}%
}
\end{table*}

\paragraph{Comparison with naive baselines.}
Tab.~\ref{tab:naive_baselines} compares TrustCLIP against Gaussian noise 
at varying intensities, evaluated on 500 SUN397 test samples under the same 
fixed attacker. The results expose a fundamental limitation of indiscriminate 
perturbation: at low noise ($\sigma \leq 0.1$), accuracy is preserved but 
privacy barely improves over the undefended baseline (DSIM: 0.33 vs.\ 0.34). 
At high noise ($\sigma = 1.0$), accuracy collapses to 17.4\% yet DSIM 
reaches only 0.68---still well below TrustCLIP. In contrast, TrustCLIP 
maintains accuracy above the baseline (85.1\%) while achieving DSIM of 0.88,
confirming that learned, selective suppression is fundamentally superior to
indiscriminate perturbation.
To rule out that this gain is merely an artefact of perturbation
magnitude, we add a control that matches the noise to TrustCLIP at equal
feature displacement: isotropic Gaussian noise calibrated to the same
$\ell_2$ perturbation $\|\tilde{z}-z\|_2$ as our projection. At this matched
budget the Gaussian control reaches only DSIM~0.64 at 12.5\% Top-1, whereas
TrustCLIP attains DSIM~0.80 at 82.9\% (undefended CLIP: 0.21 / 83.4\%)---strictly
worse on both privacy and utility. The advantage of TrustCLIP is thus a
property of \emph{where} information is removed, not \emph{how much}.
% \begin{table}[t]
% \centering
% \caption{\textbf{Comparison with naive baselines on SUN397.} 
% All methods evaluated under the same fixed attacker (trained on 
% unprotected CLIP features) on 500 shared test samples. 
% TrustCLIP achieves substantially stronger privacy at comparable 
% or higher accuracy than any noise level.}
% \label{tab:naive_baselines}
% \resizebox{\columnwidth}{!}{%
% \begin{tabular}{l c c c}
% \toprule
% Method & Top-1 & LPIPS $\uparrow$ & DSIM $\uparrow$ \\
% \midrule
% CLIP (no defense)            & 83.9 & 0.58 & 0.34 \\
% \midrule
% Noise $\sigma{=}0.01$       & 84.1 & 0.59 & 0.34 \\
% Noise $\sigma{=}0.05$       & 83.8 & 0.58 & 0.33 \\
% Noise $\sigma{=}0.1$        & 82.7 & 0.58 & 0.33 \\
% Noise $\sigma{=}0.25$       & 75.2 & 0.60 & 0.36 \\
% Noise $\sigma{=}0.5$        & 54.4 & 0.64 & 0.44 \\
% Noise $\sigma{=}1.0$        & 17.4 & 0.71 & 0.68 \\
% \midrule
% \textbf{TrustCLIP} (ours)   & \textbf{85.1} & \textbf{0.90} & \textbf{0.88} \\
% \bottomrule
% \end{tabular}%
% }
% \end{table}
\begin{table}[t]
\centering
\caption{
\textbf{Comparison with naive baselines on SUN397.} 
All methods evaluated under the same fixed attacker (trained on 
unprotected CLIP features) on test samples. TrustCLIP achieves substantially stronger privacy at comparable or higher accuracy than any noise level.}
\label{tab:naive_baselines}
\scriptsize
\setlength{\tabcolsep}{3pt}      % default ~6pt
\renewcommand{\arraystretch}{0.92} % default 1.0
\begin{tabular}{lccc}
\toprule
Method & Top-1 & LPIPS $\uparrow$ & DSIM $\uparrow$ \\
\midrule
CLIP (no defense)        & 83.9 & 0.58 & 0.34 \\
Noise $\sigma{=}0.01$    & 84.1 & 0.59 & 0.34 \\
Noise $\sigma{=}0.05$    & 83.8 & 0.58 & 0.33 \\
Noise $\sigma{=}0.1$     & 82.7 & 0.58 & 0.33 \\
Noise $\sigma{=}0.25$    & 75.2 & 0.60 & 0.36 \\
Noise $\sigma{=}0.5$     & 54.4 & 0.64 & 0.44 \\
Noise $\sigma{=}1.0$     & 17.4 & 0.71 & 0.68 \\
\midrule
\textbf{TrustCLIP} (ours) & \textbf{85.1} & \textbf{0.90} & \textbf{0.88} \\
\bottomrule
\end{tabular}
\end{table}
% \vspace{-1cm}

\paragraph{Vision-language modeling.}
Tab.~\ref{tab:vlm_unified} (top) reports TrustLLaVA alongside strong VLMs 
on the standard benchmark suite, with reconstruction metrics in the lower 
panel. TrustLLaVA preserves competitive performance on coarse-grained tasks: 
POPE (86.1 vs.\ 86.6 for LLaVA-SP) and VQAv2 (76.3 vs.\ 79.2), and notably 
\emph{improves} on LLaVA-Bench (+3.6) and MM-Vet (+0.7), suggesting that 
adversarially shaping the representation can reduce hallucination-style errors.
On the privacy side, DSIM improves from 0.32 (LLaVA-SP) to 0.39, with 
corresponding gains in LPIPS and reductions in PSNR and SSIM.
Tab.~\ref{tab:benchmark_breakdown} breaks down MME-Perception and SEED-Bench 
by category type. Tasks relying on high-level semantics---existence detection, 
artwork identification, action recognition---are fully preserved (avg.\ 
$\Delta$: --1.9\% on MME, --0.2\% on SEED). Tasks depending on fine-grained 
visual detail---counting, OCR, localization---show larger degradation (avg.\ 
$\Delta$: --11.1\% on MME, --6.4\% on SEED). This pattern directly confirms 
the premise of \S\ref{subsec:premise}: the projection selectively suppresses 
instance-specific cues while retaining the semantic structure that drives most 
VLM capabilities. The privacy gains justify these modest costs on fine-grained 
tasks, while the categories most relevant to typical VLM usage remain intact.

\begin{table*}[t!]
\centering
\caption{%
  \textbf{VLM utility and privacy evaluation.}
  \textit{Top:} Comparison with SoTA methods.
  % * indicates reproduced results using LoRA; 
  % \dag\ denotes full-training results reported in LLaVA-1.5~\cite{llava1.5}.
  % Best and second-best results are \textbf{bolded} and \underline{underlined}.
  \textit{Bottom:} Reconstruction quality when each method is attacked by 
  a dedicated IP-Adapter finetuned on its own feature distribution
  (i.e., CLIP features for LLaVA-SP, TrustCLIP features for TrustLLaVA).
  Lower PSNR/SSIM and higher LPIPS/DSIM indicate stronger privacy.
}
\label{tab:vlm_unified}

\resizebox{\textwidth}{!}{%
\begin{tabular}{l l c | c c c c | c c c c c}
\toprule
Method & LLM & Res. 
  & VQA\textsuperscript{v2} & GQA & SQA\textsuperscript{I} & VQA\textsuperscript{T} 
  & POPE & MME\textsuperscript{P} & SEED\textsuperscript{I} & LLaVA\textsuperscript{W} & MM-Vet \\
\midrule
Qwen-VL~\cite{Qwen-VL}           & Qwen-7B   & 448 & \underline{78.8} & 59.3 & 67.1 & \textbf{63.8} & --   & --             & 56.3          & --             & --   \\
Qwen-VL-Chat~\cite{Qwen-VL}      & Qwen-7B   & 448 & 78.2             & 57.5 & \underline{68.2} & \underline{61.5} & --   & \underline{1487.5} & 58.2          & --             & --   \\
LLaVA-1.5~\cite{llava1.5}    & Vicuna-7B & 336 & 78.5             & \underline{62.0} & 66.8 & 58.2 & 85.9 & \textbf{1510.7}    & 66.2          & 63.4           & 30.5 \\
LLaVA-1.5~\cite{llava1.5}       & Vicuna-7B & 336 & 78.4             & 61.9 & 67.6 & 56.2 & 85.8 & 1477.4              & \underline{67.0} & \underline{64.2} & 32.1 \\
LLaVA-SP~\cite{llava-sp}         & Vicuna-7B & 336 & \textbf{79.2}    & \textbf{62.7} & \textbf{68.4} & 58.5 & \textbf{86.6} & 1470.7 & \textbf{67.9} & 61.7 & \underline{33.8} \\
\midrule
TrustLLaVA                        & Vicuna-7B & 336 & 76.3             & 58.0 & 62.1 & 55.0 & \underline{86.1} & 1390.8 & 63.0 & \textbf{65.3} & \textbf{34.5} \\
TrustLLaVA (MLP)                   & Vicuna-7B & 336 & 72.8             & 55.4 & 60.5 & 51.1 & 85.3 & 1296.8 & 58.2 & 51.5 & 27.8 \\
\bottomrule
\end{tabular}%
}

\vspace{6pt}
\scalebox{0.9}{%
\begin{tabular}{l r r r r}
\toprule
 & \multicolumn{4}{c}{Reconstruction metrics (mean $\pm$ std)} \\
\cmidrule(lr){2-5}
 & PSNR $\downarrow$ & SSIM $\downarrow$ & LPIPS $\uparrow$ & DSIM $\uparrow$ \\
\midrule
LLaVA-SP~\cite{llava-sp}     
  & 10.57 $\pm$ 1.83 & 0.26 $\pm$ 0.13 & 0.68 $\pm$ 0.06 & 0.32 $\pm$ 0.08 \\
TrustLLaVA
  & \underline{10.42} $\pm$ 1.72 & \underline{0.24} $\pm$ 0.12 & \underline{0.70} $\pm$ 0.06 & \underline{0.39} $\pm$ 0.09 \\
TrustLLaVA (MLP)
  & \textbf{7.13} $\pm$ 1.22 & \textbf{0.12} $\pm$ 0.08 & \textbf{0.81} $\pm$ 0.06 & \textbf{0.62} $\pm$ 0.10 \\
\bottomrule
\end{tabular}%
}
\end{table*}
\begin{table}[t]
\centering
\caption{%
  \textbf{Per-category analysis of TrustLLaVA vs.\ LLaVA-SP.}
  Semantic tasks are preserved while fine-grained tasks degrade,
  confirming selective suppression of instance-specific 
  detail (\S\ref{subsec:premise}).
}
\label{tab:benchmark_breakdown}
\scalebox{0.78}{%
\begin{tabular}{l r l}
\toprule
 & Avg.\ $\Delta$ & Top categories \\
\midrule
\multicolumn{3}{l}{\textit{MME-Perception}} \\
\quad Semantic      & \textbf{--1.9\%} & Exist.\ (0.0), Art.\ (0.0), Pos.\ (--2.5) \\
\quad Fine-grained  & --11.1\%          & Post.\ (--18.9), Count (--14.5), OCR (--12.2) \\
\midrule
\multicolumn{3}{l}{\textit{SEED-Bench}} \\
\quad Semantic      & \textbf{--0.2\%} & Act.P\,(+2.9), Act.R\,(+0.9), Inter.\,(0.0) \\
\quad Fine-grained  & --6.4\%           & Count (--9.0), Loc.\ (--7.5), ID (--6.6) \\
\bottomrule
\end{tabular}%
}
\end{table}

% ==============================================================
% §4.3 PRIVACY EVALUATION
% RATIONALE: With utility established, present the privacy evidence.
% Explicitly separate fixed vs. adaptive attacker results — this was
% Reviewer z1yh's #1 confusion point, and the meta-review highlighted it.
% SOURCES:
%   - Table 1 (fixed attacker metrics during training, used for eval)
%   - Table 2 (VLM recon metrics)
%   - Supplementary §9.4 adaptive numbers (PSNR 10.62 → 11.04)
% ==============================================================
\begin{figure*}[t]
 \centering
 \includegraphics[width=0.95\textwidth]{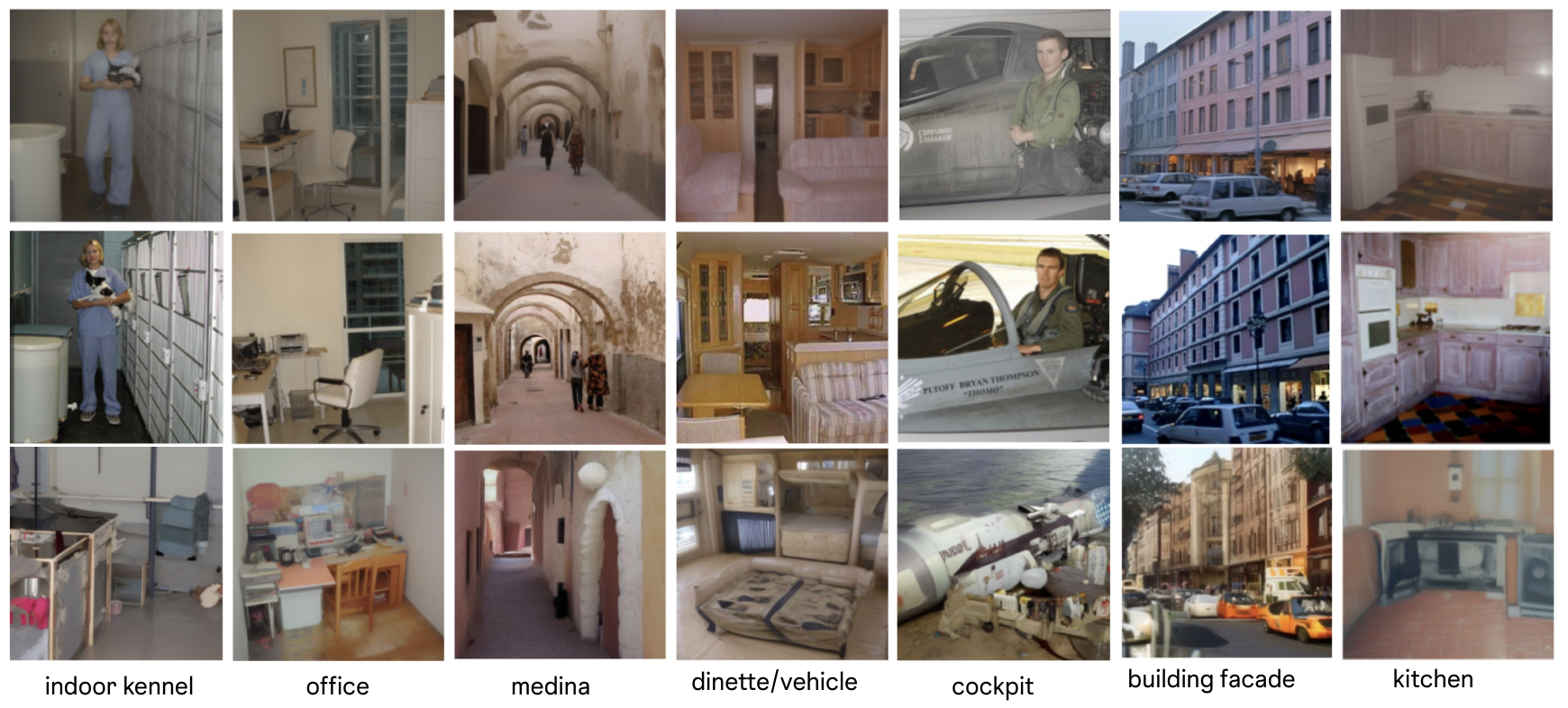}
 \vspace{-2mm}
 \caption{Qualitative comparison on SUN397~\cite{Xiao2010SUNDL}: \textbf{original} (top), reconstructions from the \textbf{vanilla CLIP} attacker (middle), and from the \textbf{TrustCLIP} attacker (bottom) under the adaptive threat model (\S\ref{subsec:threat-model}). Vanilla CLIP reveals faces, pets, textures, and distinctive color patterns; TrustCLIP obfuscates these while preserving scene semantics and class-level structure.}
 \label{fig:quals}
\end{figure*}
% \subsection{Privacy Evaluation}
% \label{subsec:privacy}
\paragraph{Robustness under adaptive and unseen attackers.}
To test the strongest threat model, we retrain a new IP-Adapter directly on
TrustCLIP features (Eq.~\ref{eq:adaptive-attacker}), giving the adversary full
knowledge of the defense. On SUN397 this adaptive attacker recovers only
marginally more detail than the fixed variant (PSNR: 11.04 vs.\ 10.62) and
remains far less effective than the vanilla CLIP attacker on unprotected
features (PSNR: 13.58); for VLMs, \suppref{Appendix~\ref{sec:supp_ablations}}{the
supplementary material} confirms consistent gains (DSIM: 0.42--0.43 for
identity-init, 0.59--0.62 for MLP, vs.\ 0.32 baseline).
The defense also holds against attackers it never co-trained with: a
higher-capacity \emph{IP-Adapter Plus} transfer attacker recovers \emph{less}
detail than our adaptive attacker (DSIM 0.88 vs.\ 0.80), and a non-diffusion
\emph{CNN decoder} fails likewise (PSNR 13.77 / DSIM 0.45 vs.\ 14.40 / 0.36 on
unprotected CLIP). Consistent failure across families---diffusion and
feed-forward alike---indicates TrustCLIP reduces the recoverable information
itself rather than overfitting one inverter (\S\ref{subsec:attacker});
see \suppref{Appendix~\ref{sec:supp_attacker_families}}{the supplementary material}.

% \input{tables/privacy-llm}

% \paragraph{Robustness under adaptive attacker.}
% To test the strongest threat model, we retrain a new IP-Adapter directly on TrustCLIP features (Eq.~\ref{eq:adaptive-attacker}), giving the adversary full knowledge of the defense.
% On SUN397, the adaptive attacker recovers marginally more detail than the fixed variant (PSNR: 11.04 vs.\ 10.62), confirming that it successfully adapts to the protected distribution.
% However, it remains far less effective than the vanilla CLIP attacker on unprotected features (PSNR: 13.58).
% The gap between the adaptive result (11.04) and the unprotected baseline (13.58) demonstrates that TrustCLIP's representations substantially limit recoverable information even under worst-case adaptation.
% For VLMs, in our supplement we report reconstruction metrics under the adaptive attacker for all identity-initialized and MLP configurations, confirming consistent privacy gains across architectures.
% % (DSIM: 0.42--0.43 for identity-init, 0.59--0.62 for MLP, vs.\ 0.32 baseline).

% ==============================================================
% §4.4 ABLATION STUDIES
% Also addresses Reviewer SUkP's joint-vs-two-stage concern.
% ==============================================================

\subsection{Ablation Studies}
\label{subsec:ablations}

\paragraph{Identity initialization vs.\ standard MLP.}
We compare two projection architectures on VLM benchmarks: 
(i)~identity-initialized projection (default) and (ii)~standard MLP without 
identity initialization. The identity-initialized variant maintains 
near-baseline performance (MM-Vet: 32.3--34.2 vs.\ 34.0 for LLaVA-SP; POPE: 
86.2--86.6 vs.\ 86.6) with meaningful privacy gains (DSIM: 0.42--0.43 vs.\ 
0.32). The MLP variant achieves stronger privacy (DSIM: 0.59--0.62) at 
significant utility cost (MM-Vet: 26.5--27.8). These configurations span a 
controllable privacy--utility spectrum, and the comparison highlights the 
importance of starting from the CLIP manifold: the identity-initialized 
variant allows task and privacy gradients to co-adapt gradually, whereas the 
MLP variant departs early from the original representation and struggles to 
recover task-relevant structure. 
% Full results in our Supplementary 
% Tables~\ref{tab:trustllava_ablation}--\ref{tab:reconstruction_metrics}.

\paragraph{Loss formulation and $\lambda_{\text{rec}}$.}
Tab.~\ref{tab:utility_privacy} ablates pixel loss ($\ell_1$ vs.\ $\ell_2$), 
perceptual loss (with/without LPIPS), and $\lambda_{\text{rec}} \in \{0.25, 
0.5, 1.0\}$ on SUN397. L2+LPIPS at $\lambda_{\text{rec}}{=}0.25$ yields the 
best trade-off: Top-1 drops 0.5\% while DSIM improves 2.5$\times$. For VLMs, 
$\lambda_{\text{rec}}{=}0.001$ suffices; performance remains stable across
three orders of magnitude (\suppref{Appendix~\ref{sec:lambda_ablation}}{supplementary material}).

\paragraph{Hyperparameters.}
For the identity-initialized projection, residual weight $r$ and 
initialization scale $\varepsilon$ have minimal impact: DSIM varies by only 
${\sim}$0.01 across configurations, with $r{=}1.0$, $\varepsilon{=}0.1$
achieving the best trade-off (\suppref{Appendix~\ref{sec:arch_comparison}}{supplementary
material tables}). For
the MLP variant, freeze duration and warmup length minimally affect privacy
(DSIM: 0.59--0.62) but modestly affect utility
(\suppref{Appendix~\ref{sec:mlp_ablation}}{Supplementary Tables}).

\begin{figure*}[t]
    \centering
    \includegraphics[width=\textwidth]{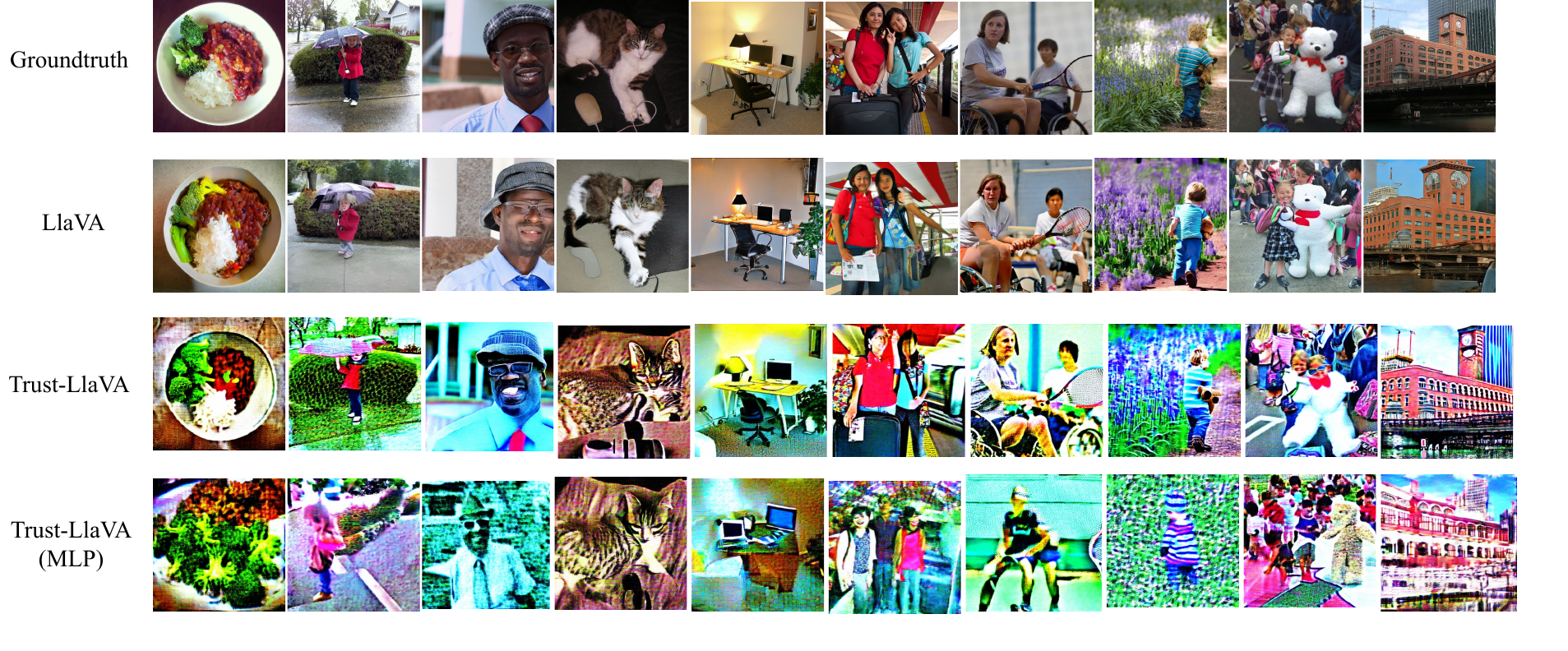}
    \caption{\textbf{Privacy spectrum on VLMs.}
    Each column shows the same COCO validation image across four settings.
    \textbf{Row~1:} Ground truth.
    \textbf{Row~2 (LLaVA):} Reconstructions from unprotected CLIP features---faces, identities, objects, and scene details are recovered with high fidelity.
    \textbf{Row~3 (Trust-LLaVA):} Identity-initialized projection---facial features become unrecognizable and fine detail is suppressed, while scene-level semantics are preserved.
    \textbf{Row~4 (Trust-LLaVA MLP):} MLP projection without identity initialization---all personally identifiable information is destroyed at the cost of reduced VLM utility (\S\ref{subsec:ablations}).
    All reconstructions use the adaptive attacker (\S\ref{subsec:threat-model}).}
    \label{fig:vlm_quals}
\end{figure*}

\subsection{Qualitative Results}
\label{subsec:qualitative}

% \paragraph{Privacy spectrum on classification.}
% Fig.~\ref{fig:lbd_priv_qual} shows the effect of $\lambda_{\text{rec}}$ on reconstructions from SUN397.
% As $\lambda$ increases from 0.25 to 1.0, reconstructions become progressively blurred: fine textures and object boundaries are lost while global scene layout remains intact, directly mirroring the quantitative trend in Tab.~\ref{tab:utility_privacy}.

% Fig.~\ref{fig:quals} extends this comparison across diverse scene categories.
Fig.~\ref{fig:quals} compares reconstructions across diverse SUN397 scene 
categories under the adaptive attacker. 
Vanilla CLIP reveals sensitive details---recognizable faces and clothing in \emph{indoor kennel} and \emph{cockpit}, vivid textures and car shapes in \emph{building facade}, and distinctive color patterns in \emph{kitchen}.
TrustCLIP systematically obfuscates these attributes: humans are reduced to vague silhouettes, facade textures are smoothed into coarse blobs, and fine-grained details are washed out.
Critically, scene semantics remain recognizable---the corridor geometry in \emph{medina}, furniture layout in \emph{dinette}, and room structure in \emph{kitchen}---confirming that the projection preserves the task-relevant information identified in \S\ref{subsec:premise}.

\paragraph{VLM reconstructions and privacy spectrum.}
Fig.~\ref{fig:vlm_quals} compares reconstructions across the full privacy spectrum.
LLaVA-SP's unprotected CLIP features (Row~2) enable high-fidelity recovery: 
the man's face and clothing are clearly recognizable, 
the cat's fur texture is preserved, group compositions and individual identities in the outdoor scenes remain distinguishable, and indoor layouts are faithfully reproduced.
The identity-initialized TrustLLaVA projection (Row~3) suppresses these 
sensitive attributes while coarse scene structure is preserved: the 
food plate, garden scene, office layout, and group configurations 
remain identifiable at the category level (DSIM: 0.42--0.43, MM-Vet: 32.3--34.2).
The MLP variant without identity initialization (Row~4) pushes privacy 
to the extreme: faces are obliterated, colors become arbitrary, and personal 
items dissolve into abstract patterns, leaving only coarse spatial layout intact (DSIM: 0.59--0.62, MM-Vet: 26.5--27.8).
Together, these configurations illustrate the controllable 
privacy--utility trade-off that TrustCLIP enables.

\section{Conclusion}
We presented TrustCLIP, a framework that strengthens the privacy of 
vision--language representations by adversarially training a lightweight 
projection to suppress the visual detail that enables generative inversion 
while preserving the semantic content needed for downstream tasks.
A central observation is that existing privacy defenses focus on discriminative 
leakage, overlooking the threat posed by diffusion-based attackers that can 
reconstruct visually coherent images from encoder features. TrustCLIP directly 
addresses this gap by optimizing against a generative reconstruction attacker 
as the primary privacy objective.
Experiments across image classification and VLM pipelines confirm that the 
approach consistently reduces inversion fidelity while maintaining competitive 
downstream performance, offering a practical privacy--utility trade-off.

\noindent\textit{Limitations and future work.}
Our evaluation focuses on CLIP encoders; while we evaluate against transfer
and non-diffusion attackers (\S\ref{subsec:main-results}), the projection is
\emph{trained} against a single attacker family (IP-Adapter + Stable
Diffusion). Extending the framework to other vision
backbones (e.g., SigLIP, DINOv2) and generative architectures is a natural
next step. Privacy suppression has a greater impact on tasks requiring
fine-grained visual detail (e.g., OCR, counting) than on coarse-grained
semantic tasks; adaptive mechanisms that modulate suppression strength
per-token or per-task could mitigate this. Finally, applying TrustCLIP to
video and multi-view settings, where temporal consistency introduces
additional privacy surfaces, is a promising direction.
\clearpage
% {
%     \small
\bibliographystyle{splncs04}
\bibliography{main}
    
% \section*{Acknowledgements}
% Please insert your acknowledgments here.

% Supplement is bound only when \sepappendix=0; set \sepappendix=1 (preamble)
% to build the main paper alone (cross-refs fall back to plain text).
\if\sepappendix1\else
% ============================================================
% SUPPLEMENTARY MATERIAL — TrustCLIP (ECCV 2026)
% ============================================================
% Place this AFTER the main paper body (remove the main \end{document}).
%
% \sepappendix flag controls cross-refs:
%   0 = joined build (arXiv): uses \ref to main paper labels
%   1 = standalone supmat PDF: uses hardcoded strings
% ============================================================

% \sepappendix and \mainref/\suppref are provided by the main document.
% Fallbacks below let this file also compile inside a standalone wrapper.
\ifdefined\sepappendix\else\def\sepappendix{0}\fi
\providecommand{\mainref}[2]{\if\sepappendix1 #2\else #1\fi}
\providecommand{\suppref}[2]{\if\sepappendix1 #2\else #1\fi}

\if\sepappendix1\else
  % Bound to the main paper: switch to appendix-style numbering.
  \appendix
  \renewcommand{\thefigure}{A.\arabic{figure}}
  \renewcommand{\thetable}{A.\arabic{table}}
  \renewcommand{\theequation}{A.\arabic{equation}}
\fi
\setcounter{figure}{0}
\setcounter{table}{0}
\setcounter{equation}{0}

% ============================================================
\section{Overview}
\label{sec:supp_intro}
% ============================================================

TrustCLIP preserves the privacy of vision--language representations while maintaining strong performance across diverse benchmarks, as illustrated in \mainref{Figure~\ref{fig:motivation}}{Figure~1}.

\noindent\textbf{Interactive results.}
We provide a browsable gallery of reconstruction samples comparing the baseline (no projection), the identity-initialized projection, and the MLP-based projection at 
\href{https://atnikos.github.io/trustclip/}{\texttt{https://atnikos.github.io/trustclip/}}.

Readers can inspect original--reconstruction pairs across all three settings, jump to specific sample indices, and visually assess the privacy--utility trade-off at scale beyond the curated examples in this document.

This supplementary material provides extended technical details, additional experiments, and further analyses that complement the main paper.
Section~\ref{sec:supp_architecture} describes the architectures of our privacy-preserving model and the generative attacker, while Section~\ref{sec:supp_implementation} outlines the training procedures for both privacy-preserving classification and VLM settings.
Section~\ref{sec:supp_ablations} presents additional ablations analyzing the privacy--utility trade-off, the effect of adaptive attacker training, and the VLM benchmark results.
Section~\ref{sec:supp_qualitative} provides qualitative examples that extend \mainref{Figures~\ref{fig:quals}--\ref{fig:vlm_quals}}{Figures~4--5} of the main paper, and Section~\ref{sec:supp_limitations} discusses failure cases and limitations.
Finally, Section~\ref{sec:supp_benchmarks} details the evaluation benchmarks corresponding to \mainref{Tables~\ref{tab:utility_privacy}--\ref{tab:vlm_unified}}{Tables~1--3} of the main paper.

% TODO: uncomment when video is ready
% \noindent\textbf{Supplementary video.}
% Our supplementary video offers an overview of the method along with additional demonstrations, including: (i)~reconstructions comparing vanilla CLIP and TrustCLIP features, (ii)~explanation of our models and adversarial losses, (iii)~results on VLM benchmarks and image classification examples.

% ============================================================
% TABLES — inlined with corrected captions
% ============================================================

% --- Table A.1: Lambda ablation (VLM) ---
\begin{table}[t]
\centering
\small
\caption{\textbf{Adversarial loss weight ablation.} We evaluate TrustLLaVA across different values of $\lambda_{\text{rec}}$, the adversarial reconstruction loss weight that controls the privacy-utility tradeoff (\mainref{Equation~\ref{eq:adv-objective}}{Equation~2} in the main paper). Lower $\lambda$ values preserve stronger task performance across all benchmarks while still providing privacy protection, with $\lambda = 0.001$ achieving the best balance. Performance remains remarkably stable across the range, demonstrating robustness to this hyperparameter. All models use identity-initialized projection with gradual warmup (\mainref{\S\ref{subsec:adv-objective}}{Section~3.3} of the main paper).}
\label{table:lambda_ablation}
\resizebox{\columnwidth}{!}{%
\begin{tabular}{l c c | c c c c c c c c c}
\toprule
Method & LLM & Res. & VQA$^{\text{v2}}$ & GQA & SQA$^{\text{I}}$ & VQA$^{\text{T}}$ & POPE & MME$^{\text{P}}$ & SEED$^{\text{I}}$ & LLaVA$^{\text{W}}$ & MM-Vet \\
\midrule
TrustLLaVA ($\lambda = 0.1$)   & Vicuna-7B & 336 & 76.5 & 58.0 & 61.7 & 54.5 & 85.5 & 1372.0 & 63.6 & 60.4 & 33.1 \\
TrustLLaVA ($\lambda = 0.01$)  & Vicuna-7B & 336 & 76.4 & 57.8 & 62.3 & 54.6 & 85.7 & 1363.6 & 63.4 & 61.2 & 33.8 \\
TrustLLaVA ($\lambda = 0.001$) & Vicuna-7B & 336 & 76.5 & 58.1 & 62.2 & 54.5 & 84.9 & 1361.0 & 63.7 & 59.4 & 34.0 \\
\bottomrule
\end{tabular}%
}
\end{table}

% --- Table A.2: Identity-init hyperparameter ablation (utility) ---
\begin{table}[t]
\centering
\small
\caption{\textbf{Identity-initialized projection hyperparameter ablation.} We evaluate TrustLLaVA performance across different configurations of the identity-initialized privacy module, varying residual weight $r \in \{0.95, 1.0\}$ and initialization scale $\varepsilon \in \{0.01, 0.1\}$. The configuration $r = 1.0$, $\varepsilon = 0.1$ achieves the best overall performance (MM-Vet: 34.2; MME$^{\text{P}}$: 1410.1). All models use $\lambda_{\text{rec}} = 0.001$ with the training schedule described in Section~\ref{sec:supp_vlm_training}.}
\label{tab:trustllava_ablation}
\resizebox{\columnwidth}{!}{%
\begin{tabular}{l c c | c c c c c c c c c}
\toprule
Method & LLM & Res. & VQA$^{\text{v2}}$ & GQA & SQA$^{\text{I}}$ & VQA$^{\text{T}}$ & POPE & MME$^{\text{P}}$ & SEED$^{\text{I}}$ & LLaVA$^{\text{W}}$ & MM-Vet \\
\midrule
TrustLLaVA ($r{=}0.95$, $\varepsilon{=}0.01$) & Vicuna-7B & 336 & 77.3 & 59.1 & 59.9 & 55.7 & 86.2 & 1391.5 & 64.5 & 62.6 & 32.3 \\
TrustLLaVA ($r{=}0.95$, $\varepsilon{=}0.1$)  & Vicuna-7B & 336 & 77.2 & 59.1 & 61.5 & 55.4 & 86.3 & 1378.5 & 64.8 & 64.1 & 32.8 \\
TrustLLaVA ($r{=}1.0$, $\varepsilon{=}0.01$)  & Vicuna-7B & 336 & 77.3 & 59.3 & 61.6 & 56.0 & 86.3 & 1394.0 & 64.6 & 62.4 & 33.1 \\
TrustLLaVA ($r{=}1.0$, $\varepsilon{=}0.1$)   & Vicuna-7B & 336 & 77.1 & 58.9 & 62.0 & 55.4 & 86.6 & 1410.1 & 64.4 & 61.5 & 34.2 \\
\bottomrule
\end{tabular}%
}
\end{table}

% --- Table A.3: MLP schedule ablation (utility) ---
\begin{table}[t]
\centering
\small
%%% FIX 3: "3-layer" → "two-layer" to match the architecture (two Linear layers) %%%
\caption{\textbf{MLP architecture training schedule ablation.} Performance of TrustLLaVA using a standard two-layer MLP projection (without identity initialization) across different adversarial training schedules. Freeze steps delay the privacy objective ($\lambda = 0$), while warmup steps gradually introduce it. All configurations show substantially degraded performance compared to identity-initialized variants (Table~\ref{tab:trustllava_ablation}), indicating that beginning from the CLIP manifold is important for VLM stability. All models use $\lambda_{\text{rec}} = 0.001$ after warmup.}
\label{tab:trustllava_schedule_ablation}
\resizebox{\columnwidth}{!}{%
\begin{tabular}{l c c | c c c c c c c c c}
\toprule
Method & LLM & Res. & VQA$^{\text{v2}}$ & GQA & SQA$^{\text{I}}$ & VQA$^{\text{T}}$ & POPE & MME$^{\text{P}}$ & SEED$^{\text{I}}$ & LLaVA$^{\text{W}}$ & MM-Vet \\
\midrule
TrustLLaVA (freeze=100, warmup=500)  & Vicuna-7B & 336 & 73.8 & 56.2 & 61.2 & 52.4 & 86.0 & 1292.2 & 60.5 & 54.1 & 27.5 \\
TrustLLaVA (freeze=100, warmup=1000) & Vicuna-7B & 336 & 73.2 & 55.9 & 63.0 & 51.8 & 84.5 & 1309.0 & 59.2 & 49.7 & 26.5 \\
TrustLLaVA (freeze=500, warmup=500)  & Vicuna-7B & 336 & 72.7 & 55.4 & 62.2 & 50.0 & 85.1 & 1269.2 & 58.4 & 50.8 & 27.8 \\
TrustLLaVA (freeze=500, warmup=1000) & Vicuna-7B & 336 & 72.8 & 55.4 & 60.5 & 51.1 & 85.3 & 1296.8 & 58.2 & 51.5 & 27.8 \\
\bottomrule
\end{tabular}%
}
\end{table}

% --- Table A.4: MLP schedule ablation (privacy metrics) ---
\begin{table}[t]
\centering
\small
\caption{\textbf{Privacy metrics for MLP architecture training schedules.} Reconstruction quality from the adaptive attacker for MLP-based TrustLLaVA variants. All configurations achieve strong privacy improvements over the LLaVA-SP baseline (DSIM: 0.59--0.62 vs.\ 0.32), with minimal variation across schedules. Lower PSNR/SSIM and higher LPIPS/DSIM indicate stronger privacy. All models use $\lambda = 0.001$ without identity initialization.}
\label{tab:reconstruction_metrics_schedule}
\centering
\resizebox{\columnwidth}{!}{%
\begin{tabular}{l | c c c c}
\toprule
 & \multicolumn{4}{c}{Reconstruction metrics (mean $\pm$ std)} \\
 & PSNR $\downarrow$ & SSIM $\downarrow$ & LPIPS $\uparrow$ & DSIM $\uparrow$ \\
\midrule
LLaVA-SP~\cite{llava-sp} & 10.57 $\pm$ 1.83 & 0.26 $\pm$ 0.13 & 0.68 $\pm$ 0.06 & 0.32 $\pm$ 0.08 \\
\midrule
TrustLLaVA (freeze=100, warmup=500)  & 7.17 $\pm$ 1.24 & 0.12 $\pm$ 0.08 & 0.80 $\pm$ 0.06 & 0.59 $\pm$ 0.10 \\
TrustLLaVA (freeze=100, warmup=1000) & 7.22 $\pm$ 1.27 & 0.12 $\pm$ 0.09 & 0.80 $\pm$ 0.06 & 0.61 $\pm$ 0.09 \\
TrustLLaVA (freeze=500, warmup=500)  & 7.13 $\pm$ 1.22 & 0.11 $\pm$ 0.08 & 0.81 $\pm$ 0.06 & 0.62 $\pm$ 0.10 \\
TrustLLaVA (freeze=500, warmup=1000) & 7.13 $\pm$ 1.22 & 0.12 $\pm$ 0.08 & 0.81 $\pm$ 0.06 & 0.62 $\pm$ 0.10 \\
\bottomrule
\end{tabular}
}
\end{table}

% --- Table A.5: Identity-init (privacy metrics) ---
\begin{table}[t]
\centering
\small
\caption{\textbf{Reconstruction quality of the diffusion-model attacker under different TrustLLaVA configurations.} Lower PSNR/SSIM and higher LPIPS/DSIM indicate stronger privacy. All configurations use $\lambda{=}0.001$.}
\label{tab:reconstruction_metrics}
\centering
\begin{tabular}{l | c c c c}
\toprule
 & \multicolumn{4}{c}{Reconstruction metrics (mean $\pm$ std)} \\
 & PSNR $\downarrow$ & SSIM $\downarrow$ & LPIPS $\uparrow$ & DSIM $\uparrow$ \\
\midrule
LLaVA-SP~\cite{llava-sp} & 10.57 $\pm$ 1.83 & 0.26 $\pm$ 0.13 & 0.68 $\pm$ 0.06 & 0.32 $\pm$ 0.08 \\
\midrule
TrustLLaVA ($r{=}0.95$, $\varepsilon{=}0.01$) & 8.53 $\pm$ 1.41 & 0.18 $\pm$ 0.12 & 0.73 $\pm$ 0.06 & 0.42 $\pm$ 0.08 \\
TrustLLaVA ($r{=}0.95$, $\varepsilon{=}0.1$)  & 8.38 $\pm$ 1.42 & 0.18 $\pm$ 0.11 & 0.73 $\pm$ 0.06 & 0.43 $\pm$ 0.08 \\
TrustLLaVA ($r{=}1.0$, $\varepsilon{=}0.01$)  & 8.52 $\pm$ 1.41 & 0.18 $\pm$ 0.12 & 0.73 $\pm$ 0.06 & 0.42 $\pm$ 0.08 \\
TrustLLaVA ($r{=}1.0$, $\varepsilon{=}0.1$)   & 8.35 $\pm$ 1.42 & 0.18 $\pm$ 0.12 & 0.73 $\pm$ 0.06 & 0.43 $\pm$ 0.08 \\
\bottomrule
\end{tabular}
\end{table}

% ============================================================
\section{Privacy Preserving Model Details}
\label{sec:supp_architecture}
% ============================================================

This section provides detailed specifications of the privacy projection module $P_\theta$ and the IP-Adapter-based attacker $G_\phi$ used throughout our experiments.

\subsection{Privacy Model}
\label{sec:supp_projection}

\mainref{Figure~\ref{fig:overview}}{Figure~3} in the main paper shows an overview of our adversarial learning pipeline including our privacy module. We train the utility task for image classification and vision-language modeling together with an adversarial reconstruction model. While we keep the weights of the vision encoder and reconstruction models frozen, we train the privacy module and task heads (linear probe for classification, MM projector and language model for VLM) and optimize the whole model jointly with the task utility loss and adversarial loss.

Following the projection defined in \mainref{\S\ref{subsec:projection}}{Section~3.1} of the main paper, the privacy projection modifies CLIP features via
\[
P_\theta(\mathbf{z}) = \mathbf{z} + f_\theta(\mathbf{z}),
\]
where $\mathbf{z} \in \mathbb{R}^{T \times D}$ denotes CLIP ViT-L/14 features with $T = 257$ tokens and $D = 1024$ dimensions.

\paragraph{MLP-based projection.}
Used in our classification experiments (\mainref{Table~\ref{tab:utility_privacy}}{Table~1} in the main paper), the MLP variant applies
\begin{align*}
\mathbf{h}_1 &= \mathrm{Linear}_{1024\rightarrow512}(\mathbf{z}), \\
\mathbf{h}_2 &= \mathrm{GELU}(\mathbf{h}_1), \\
\mathbf{h}_3 &= \mathrm{Dropout}(\mathbf{h}_2), \\
\mathbf{h}_4 &= \mathrm{LayerNorm}(\mathbf{h}_3), \\
f_\theta^{\mathrm{MLP}}(\mathbf{z}) &= \mathrm{Linear}_{512\rightarrow1024}(\mathbf{h}_4).
\end{align*}
\noindent meaning each token passes through a linear layer $\mathbb{R}^{1024} \rightarrow \mathbb{R}^{512}$, GELU activation, dropout ($p=0.1$), LayerNorm, and a final linear layer $\mathbb{R}^{512} \rightarrow \mathbb{R}^{1024}$.

%%% FIX 1: Identity-init equation — replace α=0.1 with r ∈ {0.95, 1.0} to match code and tables %%%
\paragraph{Identity-initialized projection.}
For VLM experiments, we employ an identity-preserving initialization:
\[
P_\theta^{\text{ident}}(\mathbf{z}) = \mathbf{z} + r\, \mathrm{MLP}_\theta(\mathbf{z}),
\]
where $r \in \{0.95, 1.0\}$ is a residual weight that controls the mixing strength of the learned perturbation. All linear layers in $\mathrm{MLP}_\theta$ are initialized with small Gaussian weights $\mathcal{N}(0, \varepsilon^2)$ with initialization scale $\varepsilon \in \{0.01, 0.1\}$, and the final layer uses $\mathcal{N}(0, (0.1\varepsilon)^2)$. Because $\mathrm{MLP}_\theta(\mathbf{z}) \approx \mathbf{0}$ at the start of training, $P_\theta(\mathbf{z}) \approx \mathbf{z}$ regardless of~$r$, ensuring that the projection begins as a near-identity mapping. As training progresses, the MLP learns meaningful perturbations and $r$ controls how aggressively they are applied: $r=1.0$ adds the learned perturbation at full scale, while $r=0.95$ slightly dampens it. Language models such as LLaVA expect CLIP embeddings lying on the original CLIP embedding manifold. Large initial perturbations destabilize alignment and degrade training, as shown in prior language modeling studies~\cite{houlsby2019parameter}. We ablate both $r$ and $\varepsilon$ in Tables~\ref{tab:trustllava_ablation} and~\ref{tab:reconstruction_metrics}.

\paragraph{Complexity.}
Both projection variants contain approximately 1.05M parameters ($<0.4\%$ of CLIP ViT-L/14's 304M), with $\sim$524K parameters per linear layer and an additional 1K parameters for LayerNorm (MLP variant only). The projection introduces $<1\%$ inference latency on an H100 GPU at batch size 32. For classification, projected features $\tilde{\mathbf{z}}$ are average-pooled and passed to a linear classifier; for VLM, the projected token sequence directly replaces CLIP features in LLaVA-SP's visual adapter.

\subsection{Attacker Model}
\label{sec:supp_attacker}

Our reconstruction attacker builds on IP-Adapter~\cite{ye2023ip-adapter} conditioned on Stable Diffusion v1.5~\cite{Rombach_2022_CVPR}, mapping CLIP features $\tilde{\mathbf{z}} \in \mathbb{R}^{257 \times 1024}$ to reconstructed images $\tilde{\mathbf{x}} = G_\phi(\tilde{\mathbf{z}})$. The attacker consists of three components: (i)~a frozen CLIP ViT-L/14 encoder producing $\mathbf{z}$ or $\tilde{\mathbf{z}}$,
(ii)~a learnable projection $W_{\mathrm{proj}} : \mathbb{R}^{1024} \rightarrow \mathbb{R}^{768}$ yielding SD-compatible features $\mathbf{h} = W_{\mathrm{proj}}\tilde{\mathbf{z}}$, and
(iii)~a Stable Diffusion U-Net conditioned on $\mathbf{h}$ via cross-attention. Figure~\ref{sup:fig:attacker_arch} presents the attacker architecture. To train the attacker model, we keep the image encoder, privacy module and U-Net decoder frozen, while fine-tuning the linear and cross-attention layers following the standard procedure of image-prompted adapters~\cite{ye2023ip-adapter}.

\begin{figure}[t]
\centering
\includegraphics[width=\linewidth]{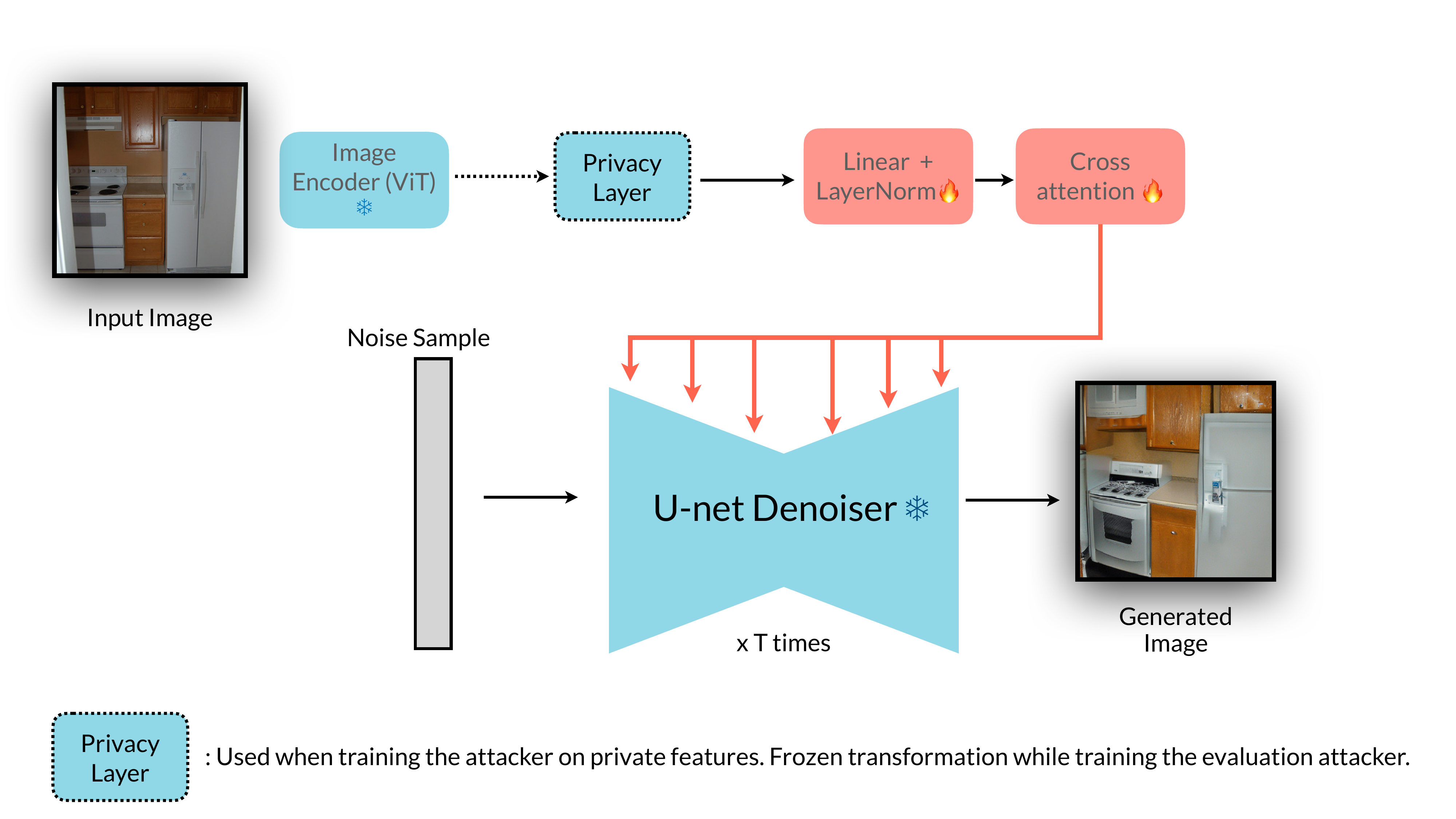}
\caption{\textbf{Adaptive reconstruction attacker based on IP-Adapter.} Visual features from the frozen CLIP encoder (with optional privacy projection for adaptive training) are mapped through trainable linear and cross-attention layers into the Stable Diffusion U-Net's conditioning space. The U-Net iteratively denoises random noise conditioned on these features to produce image reconstructions. Snowflake icons indicate frozen components; fire icons indicate trainable parameters optimized to maximize reconstruction fidelity.}
\label{sup:fig:attacker_arch}
\end{figure}

% ============================================================
\section{Training Procedures}
\label{sec:supp_implementation}
% ============================================================

\subsection{Image Classification}
\label{sec:supp_classification_training}

We evaluate TrustCLIP for classification on SUN397~\cite{Xiao2010SUNDL}, which contains 397 scene categories and 108{,}754 images. During training, the CLIP vision encoder $f_v$ is frozen while the privacy projection $P_\theta$ and classifier $h_\psi$ are optimized (see \mainref{Figure~\ref{fig:overview}}{Figure~3} in the main paper) according to:
\[
\mathcal{L}(\theta, \psi) =
\mathbb{E}_{x,y}\!\left[
\mathcal{L}_{\text{task}}^{\text{cls}}(h_\psi(\tilde{\mathbf{z}}), y)
-\lambda_{\text{rec}}\, \mathcal{L}_{\text{rec}}(x, G_\phi(\tilde{\mathbf{z}}))
\right].
\]

We use a constant-$\lambda$ schedule for classification, setting $\lambda_{\text{rec}} \in \{0.25, 0.5, 1.0\}$, without any warm-up or frozen phases; the adversarial reconstruction loss is active from the first training step. Training proceeds for 30k steps using AdamW ($\beta_1=0.9$, $\beta_2=0.999$, weight decay $0.01$), learning rate of $1\times 10^{-4}$, and batch size of 64. CLIP features are extracted at $336\times336$, while reconstructions use $512\times512$ resolution and 15 DDIM sampling steps. The reconstruction loss uses $\alpha = 0.5$, balancing L2 and LPIPS as described in \mainref{\S\ref{subsec:adv-objective}}{Section~3.3} of the main paper.

For classification, the projected tokens are average-pooled,
\[
\mathbf{u} = \frac{1}{T}\sum_{t=1}^T \tilde{\mathbf{z}}_t,
\]
and passed through a linear classifier
$\hat{\mathbf{y}} = W\mathbf{u} + \mathbf{b}$,
where $W \in \mathbb{R}^{397 \times 1024}$.

\subsection{Vision-Language Model}
\label{sec:supp_vlm_training}

To evaluate TrustCLIP in a multimodal setting, we integrate $P_\theta$ into LLaVA-SP~\cite{llava-sp} (using CLIP ViT-L/14@336px vision encoder and Vicuna-1.5-7B language model), inserting an identity-initialized projection directly before LLaVA-SP's visual adapter. We followed the standard instruction finetuning pipeline of LLaVA~\cite{llava-sp} on 665k instruction examples.

%%% FIX 2: VLM schedule — align with code: freeze=500, warmup=1000 %%%
VLM optimization follows a three-phase schedule designed to preserve multimodal alignment while gradually introducing privacy pressure:

\begin{itemize}
\item \textbf{Phase 1 (steps 0--500):} $\lambda_{\text{rec}} = 0$. Training begins without any privacy constraint to preserve the pretrained multimodal alignment and stabilize VQA learning.
\item \textbf{Phase 2 (steps 500--1500):} $\lambda_{\text{rec}}(s) = 0.001 \cdot \frac{s - 500}{1000}$. The privacy objective is gradually introduced through a linear ramp over 1{,}000 steps, allowing the model to adapt smoothly without disrupting alignment.
\item \textbf{Phase 3 (steps 1500+):} $\lambda_{\text{rec}} = 0.001$. The full privacy weight is applied for the remainder of finetuning once the model has adapted to the transition. Increasing the lambda value did not provide sufficient gains in terms of privacy and utility.
\end{itemize}

The reconstruction weight is set to a smaller value in the VLM setting ($\lambda_{\text{rec}} = 0.001$) to maintain stable multimodal optimization while the model adapts to the privacy projection. Diffusion compute is also scheduled gradually: $N_{\text{diff}} = 8$ for steps 0--3000, increased to 12 for steps 3000--6000, and kept fixed thereafter.

All remaining hyperparameters follow standard LLaVA-SP practice. We use AdamW with a learning rate of $2 \times 10^{-5}$, batch size 128 (16 per GPU across 8 H100 GPUs), LoRA rank $r=128$, and $\alpha = 0.5$ for the combined L2 + LPIPS reconstruction loss. Training requires approximately 8 hours for pretraining and 12 hours for instruction tuning.

\subsection{Attacker Training}
\label{sec:sup_attacker_training}

We finetune $W_{\mathrm{proj}}$ for 5{,}000 steps using AdamW with learning rate $1 \times 10^{-5}$ and batch size 32. To ensure our privacy evaluation is rigorous and represents a worst-case scenario, we conducted extensive ablation studies on the attacker configuration. The attacker is trained either on vanilla CLIP features $\{(\mathbf{z}_i, \mathbf{x}_i)\}$ (non-adaptive) or on TrustCLIP features $\{(\tilde{\mathbf{z}}_i, \mathbf{x}_i)\}$ (adaptive, following \mainref{Equation~\ref{eq:adaptive-attacker}}{Equation~1} in the main paper).

We systematically explored a comprehensive range of hyperparameters to maximize reconstruction quality: we varied attacker capacity (testing projection layers with 1--8 blocks and hidden dimensions from 768 to 3072), extended training horizons (from 5k to 50k steps, verifying convergence through loss plateaus), adjusted learning rates across three orders of magnitude ($10^{-6}$ to $10^{-4}$), and experimented with different sampling parameters including guidance scales (3.0--15.0) and DDIM step counts (8--50). In all configurations, we observed clear convergence patterns, with reconstruction losses plateauing after approximately 3{,}000--4{,}000 steps and showing negligible improvements ($<0.1\%$ in LPIPS) beyond our chosen 5{,}000 step configuration. This extensive search yielded marginal gains beyond our selected parameters, providing strong evidence that the attacker operates near its theoretical capacity; additional architectural complexity, extended training, or hyperparameter tuning does not meaningfully improve inversion quality. This saturation in attack performance strengthens our privacy claims, as it suggests we are evaluating against near-optimal reconstruction adversaries rather than undertrained baselines.

All visualizations shown in the main paper and supplementary material use adaptive attackers trained on the private features each experiment yields, representing the strongest possible attack scenario. At test time, we extract CLIP features at $336 \times 336$ resolution, apply $P_\theta$ to obtain $\tilde{\mathbf{z}}$, project them into SD latent space, and generate reconstructions via DDIM sampling (guidance scale 7.5). We use $N=15$ DDIM steps for classification experiments and $N=8$--$12$ steps for VLM experiments.

% ============================================================
\section{Additional Ablation Studies and Analysis}
\label{sec:supp_ablations}
% ============================================================

We provide additional ablation studies and analysis on the adversarial loss regularization parameter, the identity-initialized projection, the training schedule, and the adaptive attacker design.

\subsection{Adversarial Loss Weight}
\label{sec:lambda_ablation}

Table~\ref{table:lambda_ablation} explores the impact of the adversarial reconstruction loss weight $\lambda_{\text{rec}}$ on VLM performance. Remarkably, performance remains stable across three orders of magnitude ($\lambda \in \{0.1, 0.01, 0.001\}$), with variations typically within 1--2 points across benchmarks. The smallest weight ($\lambda=0.001$) emerges as optimal, achieving the highest MM-Vet score (34.0) while maintaining competitive performance on other metrics. This stability suggests that even minimal adversarial pressure ($\lambda=0.001$) suffices to induce privacy-preserving features without catastrophically disrupting the pretrained vision-language alignment. The marginal performance differences between $\lambda$ values indicate that the identity-initialized architecture provides inherent robustness to the privacy objective's strength. Notably, increasing $\lambda$ to 0.1 shows slight degradation on perception tasks (MME$^{\text{P}}$: 1372.0 vs.\ 1361.0) without proportional privacy gains, confirming that $\lambda=0.001$ strikes the optimal balance for VLMs. This contrasts sharply with classification experiments where $\lambda \in \{0.25, 0.5, 1.0\}$ are viable, highlighting VLMs' greater sensitivity to representation perturbations.

\subsection{Privacy--Utility Pareto Analysis}
\label{sec:pareto}

\subsection{Identity-Initialized Projection}
\label{sec:arch_comparison}

We systematically compare two architectural variants of the privacy projection module $P_\theta$: (i)~a standard MLP without identity initialization (Tables~\ref{tab:trustllava_schedule_ablation} and~\ref{tab:reconstruction_metrics_schedule}), and (ii)~an identity-initialized MLP projection with controlled perturbation (Tables~\ref{tab:trustllava_ablation} and~\ref{tab:reconstruction_metrics}).

The MLP architecture achieves substantially stronger privacy protection, with PSNR dropping to 7.13--7.22 compared to 8.35--8.53 for identity initialization, and DSIM reaching 0.59--0.62 versus 0.42--0.43. This represents a $\sim$40\% improvement in semantic distance (DSIM) and $\sim$15\% reduction in structural similarity (PSNR). However, this enhanced privacy comes at a utility cost: MM-Vet performance plummets from 32.3--34.2 (identity) to 26.5--27.8 (MLP), while MME$^{\text{P}}$ drops from $\sim$1390 to $\sim$1290. The identity-initialized variant maintains near-baseline performance on POPE (86.2--86.6) while MLP degrades to 84.5--86.0.

The identity initialization acts as a critical anchor for preserving vision-language alignment. Without it, the model struggles to maintain multimodal coherence even with careful training schedules, suggesting that starting from the CLIP manifold is essential for VLMs.

% TODO: uncomment when MLP privacy figure is added
% Figure~\ref{fig:mlp_privacy_comparison} visually demonstrates this dramatic privacy enhancement: the MLP architecture without identity initialization produces heavily corrupted reconstructions where faces dissolve into abstract color patterns and private spaces become unrecognizable, achieving near-complete visual anonymization at the cost of utility.

\paragraph{Hyperparameter Analysis.}

For the identity-initialized architecture (Tables~\ref{tab:trustllava_ablation} and~\ref{tab:reconstruction_metrics}), we also ablate residual weight $r \in \{0.95, 1.0\}$ and initialization scale $\varepsilon \in \{0.01, 0.1\}$: Setting $r=1.0$ (full residual connection) generally outperforms $r=0.95$ on utility metrics, with MM-Vet reaching 34.2 (best overall) for $r=1.0, \varepsilon=0.1$. The privacy impact is minimal, with DSIM differing by only $\sim$0.01 between residual weights, suggesting the initialization scale $\varepsilon$ dominates privacy characteristics. Larger initialization ($\varepsilon=0.1$) provides slightly better privacy (DSIM: 0.43 vs.\ 0.42, PSNR: 8.35--8.38 vs.\ 8.52--8.53) without substantially harming utility. Interestingly, $\varepsilon=0.1$ sometimes improves performance (MM-Vet: 34.2 vs.\ 33.1 for $r=1.0$), possibly due to beneficial regularization effects. The $r=1.0, \varepsilon=0.1$ configuration emerges as the best overall, achieving the highest MM-Vet (34.2) and MME$^{\text{P}}$ (1410.1) scores while maintaining competitive privacy (DSIM=0.43).

\paragraph{Cross-Architecture Insights.}

Comparing across architectures reveals fundamental trade-offs: The MLP architecture without identity initialization achieves $\sim$45\% stronger privacy (DSIM: 0.62 vs.\ 0.43) but suffers $\sim$20\% utility degradation on average. The identity-initialized variant maintains 95\%+ of baseline performance on most benchmarks while still providing meaningful privacy (DSIM improvement from 0.32 to 0.43 over baseline LLaVA-SP). OCR-dependent tasks (VQA$^{\text{T}}$) show similar degradation patterns across architectures (50--52 for MLP, 55--56 for identity), while compositional reasoning (GQA, SQA$^{\text{I}}$) exhibits higher sensitivity to architecture choice (55--56 for MLP vs.\ 59--62 for identity). Identity initialization enables stable training with minimal hyperparameter sensitivity, while a standard MLP without identity initialization requires careful schedule tuning yet still exhibits higher variance across benchmarks.

\subsection{Training Schedule}
\label{sec:mlp_ablation}

We explore how freeze and warmup schedules affect the privacy-utility balance in Tables~\ref{tab:trustllava_schedule_ablation} and~\ref{tab:reconstruction_metrics_schedule}. Extending the freeze period from 100 to 500 steps shows mixed effects. While longer freezing slightly improves stability on some benchmarks (MM-Vet: 27.8 vs.\ 26.5--27.5 for freeze=500, warmup=1000), it generally reduces performance on reasoning tasks (GQA drops from 56.2 to 55.4). Privacy metrics remain remarkably stable across freeze periods, with DSIM varying only between 0.59--0.62, suggesting the freeze duration primarily affects utility rather than privacy. Longer warmup (1000 vs.\ 500 steps) provides marginal privacy gains (DSIM: 0.61 vs.\ 0.59 for freeze=100). The freeze=500, warmup=500 setting achieves the best privacy (DSIM=0.62, LPIPS=0.81) with reasonable utility preservation.

% TODO: uncomment when MLP privacy figure is added
% The qualitative impact of these MLP configurations is striking, as shown in Figure~\ref{fig:mlp_privacy_comparison}, where even optimally-scheduled MLP training produces psychedelic, privacy-maximizing reconstructions that obliterate personally identifiable information.

\subsection{Adaptive vs.\ Non-Adaptive Attackers}
\label{sec:adaptive_comparison}

The \textit{non-adaptive} attacker is trained on standard CLIP features,
\[
\phi^{\text{non-adap}} =
\arg\min_\phi \mathbb{E}_{x \sim \mathcal{D}}
\mathcal{L}_{\text{rec}}(x, G_\phi(f_v(x))),
\]
while the \textit{adaptive} attacker is trained directly on TrustCLIP features produced by the optimized projection $P_\theta^*$,
\[
\phi^{\text{adap}} =
\arg\min_\phi \mathbb{E}_{x \sim \mathcal{D}}
\mathcal{L}_{\text{rec}}(x, G_\phi(P_\theta^*(f_v(x)))).
\]

All qualitative results in \mainref{Figures~\ref{fig:vlm_quals}--\ref{fig:quals}}{Figures~2--4} of the main paper are generated using the adaptive attacker, representing the strongest and most informed threat model.

The adaptive attacker recovers more visual detail than the non-adaptive variant (e.g., PSNR improves from 10.62 to 11.04), indicating that our attacker is effectively trained on the TrustCLIP feature distribution. However, the adaptive attacker remains far less effective than the vanilla CLIP attacker operating on unprotected CLIP features. This gap highlights that TrustCLIP's representations substantially limit reconstructible information, providing strong privacy protection even under an adaptive threat model.

% ============================================================
\section{Generalization and Information-Level Analysis}
\label{sec:supp_generalization}
% ============================================================

This section provides the extended experiments summarized in the main paper:
generalization of the defense to attacker families not used during training
(\S\ref{sec:supp_attacker_families}), an analysis of \emph{what} information is
removed versus preserved (\S\ref{sec:supp_what_removed}), a test of the one-hot
collapse hypothesis via feature entropy (\S\ref{sec:supp_entropy}), and the
computational overhead of the privacy projection (\S\ref{sec:supp_compute}).

\subsection{Generalization to Unseen Attacker Families}
\label{sec:supp_attacker_families}

A natural concern is whether TrustCLIP merely overfits the specific IP-Adapter
checkpoint used as the training-time attacker, rather than removing
reconstructible information from the features. To test this, we evaluate the
\emph{same} frozen TrustCLIP features against two attackers that were never used
to train $P_\theta$ (Table~\ref{tab:supp_attacker_families}):
(i)~\textbf{IP-Adapter Plus}, a higher-capacity variant trained only on vanilla
CLIP features (a pure transfer attacker), and
(ii)~a \textbf{CNN decoder}, a feed-forward
inverter with no diffusion process, latent space, or cross-attention
(architecture detailed below).

The transfer IP-Adapter Plus attacker recovers \emph{less} detail from TrustCLIP
features than our specialized adaptive attacker (DSIM 0.88 vs.\ 0.80), confirming
that the defense does not depend on the exact attacker checkpoint. The
non-diffusion CNN decoder likewise fails on protected features (DSIM 0.45),
remaining close to its reference quality on unprotected CLIP (0.36) and far below
faithful reconstruction. That attackers from two distinct families---diffusion
and feed-forward convolutional---both fail is consistent with the
data-processing inequality argument of \mainref{\S\ref{subsec:attacker}}{Section~3.4}:
no decoder can recover more about $x$ than $\tilde{z}$ contains, so a defense that
reduces the information in $\tilde{z}$ degrades \emph{every} downstream inverter,
not just the one it trained against.

\begin{table}[t]
\centering
\small
\caption{\textbf{Generalization across attacker families.} The same frozen
TrustCLIP features are attacked by inverters never used to train $P_\theta$: a
higher-capacity transfer attacker (IP-Adapter Plus, trained only on vanilla CLIP)
and a non-diffusion CNN decoder. ``Ours'' = TrustCLIP. Lower PSNR and higher DSIM
indicate stronger privacy. Both unseen attackers fail to invert TrustCLIP
features, showing the defense is not specific to the training-time attacker.}
\label{tab:supp_attacker_families}
\begin{tabular}{lcc}
\toprule
Attacker family & PSNR$\downarrow$ & DSIM$\uparrow$ \\
\midrule
CLIP + IP-Adapter (reference) & 13.58 & 0.21 \\
CLIP + CNN decoder (reference) & 14.40 & 0.36 \\
\midrule
Ours + IP-Adapter (paper attacker) & 10.47 & 0.80 \\
Ours + IP-Adapter Plus (transfer) & 9.87 & 0.88 \\
Ours + CNN decoder (non-diffusion) & 13.77 & 0.45 \\
\bottomrule
\end{tabular}
\end{table}

\paragraph{CNN-decoder architecture.}
The CNN decoder is a non-diffusion attacker that reconstructs images directly
from CLIP patch embeddings $\mathbf{z} \in \mathbb{R}^{257 \times D}$. It first
drops the \texttt{CLS} token (leaving 256 patch tokens), applies a linear
projection $D \rightarrow 512$, and reshapes the result into a $16{\times}16{\times}512$
spatial feature map. Four transposed-convolution blocks then upsample this map,
reducing channels $512 \rightarrow 256 \rightarrow 128 \rightarrow 64 \rightarrow 32$
while increasing spatial resolution $16 \rightarrow 32 \rightarrow 64 \rightarrow
128 \rightarrow 256$; each block is a \texttt{ConvTranspose2d} followed by
\texttt{BatchNorm} and \texttt{GELU}. A final $3{\times}3$ convolution maps
$32 \rightarrow 3$ channels with a sigmoid activation, and the output is
bilinearly resized to $224{\times}224$, yielding an RGB image in $[0,1]$. The
decoder has ${\sim}$10M parameters and uses no skip connections, no iterative
refinement, no latent space, and no cross-attention; it is fully standalone and
does not depend on any diffusion model. We train it with the same reconstruction
objective and data as the IP-Adapter attackers.

\subsection{What Survives Reconstruction? Identity vs.\ Attributes}
\label{sec:supp_what_removed}

To characterize \emph{what} the projection removes, we run a CelebA study that
separates identity-level content (measured on reconstructions) from
attribute-level content (measured on the frozen features); see
Table~\ref{tab:supp_what_removed}. We invert features with the adaptive attacker
and analyze the outputs with off-the-shelf face tooling. On vanilla CLIP
reconstructions, InsightFace detects a face in 70.2\% of images; on TrustCLIP
reconstructions this drops to 19.8\%. Among the faces that are still detected,
the ArcFace cosine similarity to the original identity is at chance for TrustCLIP
and only marginally above chance for CLIP---both well below any operational
verification threshold. The inverter therefore recovers scene-level appearance,
not recognizable identity.

Conversely, attribute information that does not enable reconstruction is
preserved: a 4-way hair-colour linear probe on the frozen features drops only
1.5\% (91.4\%~$\rightarrow$~89.9\%). Identity-level content on reconstructions
collapses while attribute-level content on features survives---direct evidence
for the separability premise of \mainref{\S\ref{subsec:premise}}{Section~3.2}.

\begin{table}[t]
\centering
\small
\caption{\textbf{What survives reconstruction?} Identity-level signal is
measured on attacker reconstructions; attribute- and diversity-level signal is
measured on the frozen features. ``Ours'' = TrustCLIP. Identity collapses while
attributes and feature diversity are retained.}
\label{tab:supp_what_removed}
\begin{tabular}{lccc}
\toprule
What survives reconstruction? & CLIP & Ours & $\Delta$ \\
\midrule
\multicolumn{4}{l}{\emph{Identity-level (measured on reconstructions)}}\\
\quad Face detected in reconstruction (\%) & 70.2 & 19.8 & $-50.4$ \\
\midrule
\multicolumn{4}{l}{\emph{Attribute / diversity (measured on frozen features)}}\\
\quad CelebA hair-colour (4-way) probe (\%) & 91.4 & 89.9 & $-1.5$ \\
\quad Per-token entropy ($\max=\log 1280$) & 6.85 & 6.87 & $+0.3\%$ \\
\bottomrule
\end{tabular}
\end{table}

\subsection{Feature Entropy and the One-Hot Collapse Question}
\label{sec:supp_entropy}

One might worry that the projection achieves privacy trivially by collapsing
features toward a low-entropy, near one-hot code that encodes only the class
label, which would defeat reconstruction but also destroy transferable
structure. The per-token entropy in Table~\ref{tab:supp_what_removed} rules this
out: TrustCLIP preserves per-token entropy within 0.3\% of CLIP's
(6.87 vs.\ 6.85), whereas a one-hot code would have entropy near zero. This also
explains why a projection frozen after classification training transfers to VLM
tasks without re-training: the protected features retain nearly all of their
representational diversity, and the defense removes reconstruction-specific
detail rather than collapsing the representation.

\subsection{Computational Overhead}
\label{sec:supp_compute}

The privacy projection adds negligible inference cost. On an A100 at batch size
32, end-to-end inference with $P_\theta$ takes 113.5\,ms versus 111.4\,ms without
it---an overhead below 2\%---consistent with the parameter count reported in
\S\ref{sec:supp_projection} ($\sim$1.05M parameters, ${<}0.4\%$ of CLIP
ViT-L/14). Training the classification projection takes ${\sim}$24\,h on a single
H200 GPU; TrustLLaVA reuses the LLaVA-SP wall-clock plus a constant per-step
reconstruction overhead from the frozen attacker. Code and a full hyperparameter
table will accompany the release.

% ============================================================
\section{Additional Qualitative Analysis}
\label{sec:supp_qualitative}
% ============================================================

\paragraph{Privacy spectrum on classification.}
Fig.~\ref{fig:lbd_priv_qual} shows the effect of $\lambda_{\text{rec}}$ on attacker reconstructions from SUN397.
As $\lambda$ increases from 0.25 to 1.0, reconstructions become progressively blurred: fine textures and object boundaries are lost while global scene layout remains intact, directly mirroring the quantitative trend in \mainref{Tab.~\ref{tab:utility_privacy}}{Tab.~1} of the main paper.

\begin{figure}[t]
 \centering
 \includegraphics[width=0.88\textwidth]{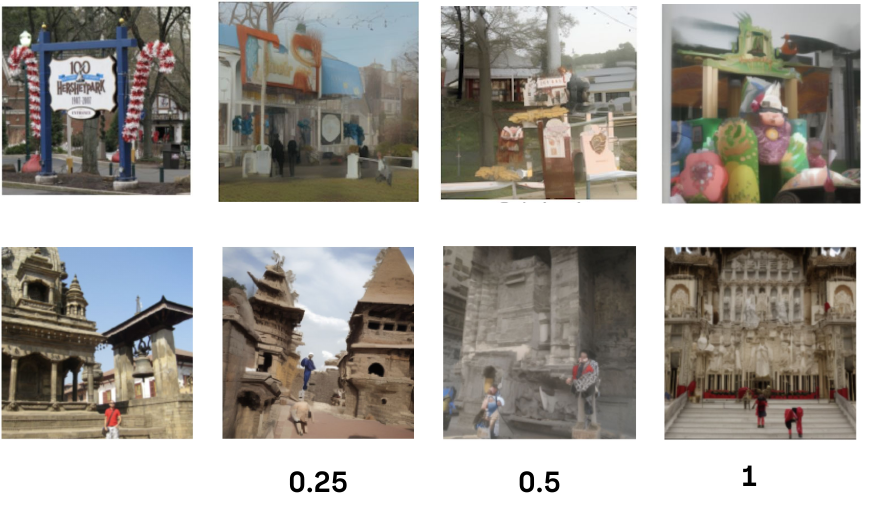}
\vspace{-2mm}
 \caption{\textbf{Effect of $\lambda_{\text{rec}}$ on reconstructions.} Left: original images. Right three columns: attacker reconstructions at $\lambda = 0.25$, $0.5$, $1.0$. Increasing $\lambda$ progressively suppresses fine-grained structure while preserving global scene layout.}
 \label{fig:lbd_priv_qual}
\end{figure}

\section{Failure Cases and Limitations}
\label{sec:supp_limitations}
% ============================================================

While TrustCLIP provides strong privacy protection across a range of settings, several trade-offs are worth noting. First, privacy constraints have a greater impact on tasks requiring fine-grained visual details (e.g., counting or text recognition) than on coarse-grained classification, which is expected given the reduced visual precision in the protected feature space. Second, stronger projection modules can further enhance privacy but may introduce additional utility loss in complex VLM tasks, making lightweight identity-initialized projections a practical balance between privacy and performance. Finally, reconstructed images under strong privacy settings still retain coarse spatial layout, which is typical for diffusion-based inversion and remains necessary for spatial reasoning tasks in VLMs. These observations highlight avenues for future exploration, including adaptive privacy mechanisms, alternative projection architectures, and the study of feature spaces from other vision backbones (e.g., SigLIP~\cite{zhai2023sigmoid}, DINOv2~\cite{oquab2023dinov2}) and their associated decoders.

% ============================================================
\section{Evaluation Benchmark Details}
\label{sec:supp_benchmarks}
% ============================================================

\subsection{Image Classification}
\label{sec:supp_sun397}

We evaluate TrustCLIP for classification on the SUN397 dataset~\cite{Xiao2010SUNDL}. SUN397 contains 397 scene categories and involves privacy-sensitive environments, including bedrooms, bathrooms, hospital rooms, children's rooms, offices, operating rooms, pharmacies, and jail cells. We report Top-1 accuracy, Top-5 accuracy, and mean class accuracy.

\subsection{Vision-Language Benchmarks}
\label{sec:supp_vlm_benchmarks}

We evaluate TrustCLIP-integrated VLMs across a broad suite of established multimodal benchmarks:

\noindent\textbf{VQA$^{\text{v2}}$}~\cite{goyal2017making}: A large-scale dataset built on 204{,}721 MS-COCO images with over 1.1M questions spanning yes/no, counting, and open-ended categories. Evaluation follows the standard human-consensus metric, where predictions are scored by agreement with crowd-sourced answers.

\noindent\textbf{GQA}~\cite{hudson2019gqa}: 113K images paired with 22M compositional reasoning questions grounded in scene graphs.

\noindent\textbf{VQA$^{\text{T}}$}~\cite{singh2019towards}: 28{,}408 images and 45{,}336 questions requiring OCR-centric reasoning---highly relevant for privacy-sensitive text content.

\noindent\textbf{ScienceQA$^{\text{I}}$}~\cite{lu2022learn}: 6{,}218 science problems involving diagrams, charts, and domain-specific visual understanding.

\noindent\textbf{MM-Vet}~\cite{yu2023mmvet}: 218 challenging examples covering recognition, OCR, knowledge, spatial reasoning, math, and language tasks, graded using GPT-4 on a 0--100 scale.

\noindent\textbf{LLaVA$^{\text{W}}$}~\cite{liu2023visualinstruction}: 24 images with 60 instruction-following questions, evaluated via GPT-4 comparative scoring.

\noindent\textbf{MME$^{\text{P}}$}~\cite{fu2023mme}: 2{,}374 images across 14 visual tasks (existence, counting, position, color, celebrity, scene, OCR), aggregated into a single performance score.

\noindent\textbf{SEED$^{\text{I}}$}~\cite{li2024seedbench}: 19{,}242 questions probing 12 dimensions, including scene, identity, attributes, counting, and spatial relationships.

\noindent\textbf{POPE}~\cite{li2023pope}: A hallucination-focused benchmark consisting of 500 images, each paired with 6 binary existence queries. Performance is measured using accuracy, F1 score, and the yes-ratio, providing a targeted assessment of object presence hallucination in vision--language models.
\fi

\end{document}